\documentclass[conference]{IEEEtran}

\ifCLASSINFOpdf
\else
\fi

\usepackage{tikz}
\usepackage{amsmath}
\usepackage{graphicx}
\usepackage{subfigure}
\usepackage{amsfonts,latexsym}
\usepackage{microtype}
\usepackage{booktabs}
\usepackage{subfigure}
\usepackage{tabularx}
\usepackage{indentfirst} 
\usepackage{multicol}
\usepackage{pdfpages}
\usepackage{multirow}
\usepackage{listings}
\usepackage{xcolor}
\usepackage{xspace}
\usepackage{url}
\usepackage{caption}
\usepackage{algorithm}
\usepackage{algpseudocode}
\usepackage{comment}
\newcommand\sysname{Scale-MIA\xspace}

\begin{document}

\title{\sysname: A Scalable Model Inversion Attack against Secure Federated Learning via Latent Space Reconstruction}

\author{\IEEEauthorblockN{Shanghao Shi\IEEEauthorrefmark{1},
Ning Wang\IEEEauthorrefmark{2},
Yang Xiao\IEEEauthorrefmark{3}, 
Chaoyu Zhang\IEEEauthorrefmark{1},
Yi Shi\IEEEauthorrefmark{1},
Y. Thomas Hou\IEEEauthorrefmark{1},
and Wenjing Lou\IEEEauthorrefmark{1}}
\IEEEauthorblockA{\IEEEauthorrefmark{1} Virginia Polytechnic Institute and State University, VA, USA \\ \{shanghaos, chaoyu, yshi, thou, wjlou\}@vt.edu}
\IEEEauthorblockA{\IEEEauthorrefmark{2} University of South Florida, FL, USA\\
{ningw}@usf.edu}
\IEEEauthorblockA{\IEEEauthorrefmark{3} University of Kentucky, KY, USA\\
{xiaoy}@uky.edu}}

\IEEEoverridecommandlockouts
\makeatletter\def\@IEEEpubidpullup{6.5\baselineskip}\makeatother
\IEEEpubid{\parbox{\columnwidth}{
		Network and Distributed System Security (NDSS) Symposium 2025\\
		24-28 February 2025, San Diego, CA, USA\\
		ISBN 979-8-9894372-8-3\\
		https://dx.doi.org/10.14722/ndss.2025.240644\\
		www.ndss-symposium.org
}
\hspace{\columnsep}\makebox[\columnwidth]{}}

\maketitle

\begin{abstract}

Federated learning is known for its capability to safeguard the participants' data privacy. However, recently emerged model inversion attacks (MIAs) have shown that a malicious parameter server can reconstruct individual users' local data samples from model updates. 
The state-of-the-art attacks either rely on computation-intensive iterative optimization methods to reconstruct each input batch, making scaling difficult, or involve the malicious parameter server adding extra modules before the global model architecture, rendering the attacks too conspicuous and easily detectable. 

To overcome these limitations, we propose \sysname, a novel MIA capable of efficiently and accurately reconstructing local training samples from the aggregated model updates, even when the system is protected by a robust secure aggregation (SA) protocol.
\sysname utilizes the inner architecture of models and identifies the latent space as the critical layer for breaching privacy. \sysname decomposes the complex reconstruction task into an innovative two-step process. The first step is to reconstruct the latent space representations (LSRs) from the aggregated model updates using a closed-form inversion mechanism, leveraging specially crafted linear layers. Then in the second step, the LSRs are fed into a fine-tuned generative decoder to reconstruct the whole input batch.

We implemented \sysname on commonly used machine learning models and conducted comprehensive experiments across various settings. The results demonstrate that \sysname achieves excellent performance on different datasets, exhibiting high reconstruction rates, accuracy, and attack efficiency on a larger scale compared to state-of-the-art MIAs. Our code is available at \url{https://github.com/unknown123489/Scale-MIA}.

\end{abstract}

\section{Introduction}

Federated learning (FL) is a distributed learning framework that enables its participants to collaboratively train a machine learning model without sharing their individual datasets \cite{mcmahan2017communication}. Within this framework, the training process occurs iteratively between a central parameter server and a group of clients. During each training round, the parameter server first broadcasts a global model with a pre-agreed model architecture to all or a fraction of clients. 
The server then collects and aggregates the model updates (either gradient or parameter updates) submitted by the clients, which are trained and derived from their respective local datasets. As no individual training data samples are exchanged between participants in this process, FL is widely recognized as a communication-efficient and privacy-preserving learning paradigm.

\subsection{Privacy Leakage of Federated Learning}

Unfortunately, recent research reveals that the privacy of FL is susceptible to breaches, enabling the attackers to infer information about the clients' proprietary datasets \cite{enthoven2021overview}. Of particular concern are the \textit{model inversion attacks (MIAs)} \cite{zhu2019deep,zhao2020idlg,geiping2020inverting,zhu2020r,yin2021see,lu2022april,hatamizadeh2022gradvit}, 
in which the adversary tries to reconstruct the original training samples from the model updates submitted by the clients. In these attacks, the parameter server is considered an \textit{honest-but-curious attacker} whose goal is to closely approximate the input samples by minimizing the distance between real gradients uploaded by individual clients and those generated by approximated dummy samples. These attacks can successfully reconstruct high-fidelity training samples when the server gains access to \textit{individual model updates} and undergoes sufficient optimization iterations. 

To counter these attacks, Bonawitz et al. propose the \textit{secure aggregation (SA)} protocol \cite{bonawitz2017practical} to prevent the server from gaining knowledge about individual model updates. SA is a specialized secure multi-party computation (MPC) protocol that allows the server to compute the summation of model updates without knowing individual values. SA ensures that individual model updates are cryptographically masked and the server cannot distinguish them from \textit{random numbers}. SA is considered one of the most robust defense mechanisms against various inference attacks targeting federated learning systems \cite{huang2021evaluating}, and several follow-up works have been proposed to further reduce the communication and computation overhead of the original SA protocol \cite{bell2020secure,guo2020v,kadhe2020fastsecagg,xu2019verifynet}.

Despite SA's initial security guarantees, recent privacy attacks show that the SA protocol is breakable when the attacker can modify the model parameters or architectures, which \textit{goes beyond the honest-but-curious threat model}.
Two distinct attack strategies have been identified to break the SA protocol. The first strategy involves obtaining individual model updates from the aggregated results by carefully manipulating the global model parameters and having the target client's model update dominate the aggregated results. This can be achieved by eliminating the model updates of all other clients except the target \cite{pasquini2022eluding}, or amplifying only the gradients of the target victim \cite{wen2022fishing}. After obtaining individual model updates, the attacker can utilize the existing optimization-based MIAs to reconstruct input samples.
However, these attacks still require costly optimization-based MIAs as part of their attack flows, limiting their applicability at scale.

The second strategy involves a more direct approach to reconstructing the input samples just from the aggregated results. To accomplish this, existing work requires the attacker to insert either a two-layer linear module \cite{fowl2021robbing} or a convolutional module \cite{zhao2023secure} before the pre-agreed global model architecture, as well as possessing a representative auxiliary dataset. These modules are meticulously crafted or trained using the auxiliary dataset, enabling the attacker to reconstruct the inputs from the gradients of crafted layers within these modules, using customized analytical methods. However, modifying the pre-agreed model architecture is highly conspicuous and is unlikely to be accepted by the clients.

\subsection{Our Attack}

In this paper, we present a novel model inversion attack named \sysname to break the secure aggregation (SA) protocol in FL. \sysname is designed to be scalable, stealthy, efficient, and highly effective, overcoming the deficiencies of existing attacks. It is capable of accurately reconstructing the clients' local data samples from the aggregated results, requires no modifications to the pre-agreed model architecture, and is more difficult to detect. \sysname also eliminates the costly per batch or sample search-based optimization process. This enables the reconstruction of hundreds of data samples in parallel and results in significant computational speed-up. 

The proposed \sysname involves two distinct phases---the \textit{adversarial model generation} phase and the actual \textit{input reconstruction} phase. 
The adversarial model generation can be done offline by the malicious parameter server once at the outset. The purpose is to generate an adversarial global model that follows the same architecture as the pre-agreed model architecture that will be distributed to the clients during the FL iterations. More specifically, the attacker first trains a surrogate autoencoder, with its encoder having \textit{the same} architecture to the encoder of the real global model, and a customized generative decoder capable of reconstructing the inputs, utilizing a collected auxiliary dataset. Subsequently, the attacker feeds the auxiliary dataset to the already-trained encoder, enabling the estimate of essential statistical parameters for crafting linear layers in the adversarial global model. 
Finally, the adversarial global model is assembled by having its encoder identical to the surrogate autoencoder, and the following linear layers are crafted with the estimated parameters.

In the second phase, the attacker disseminates the crafted adversarial global model to clients and awaits local updates from them. Assuming the presence of the SA protocol, the attacker only receives the aggregated model updates from these clients. The proposed input reconstruction phase takes the aggregated model updates as input and aims to reconstruct as many local samples as possible. We design a novel model inversion method, in which we decompose the input reconstruction phase into two steps, both only involving efficient matrix computation and feed-forward neural network computations to reduce its complexity, making it super efficient to conduct. More specifically, we first disaggregate the received aggregated model update into batched latent space representations (LSRs) through a closed-form \emph{linear leakage} module and then feed these representations into the pre-trained generative decoder to reconstruct the original input batch. This phase can be executed in a single federated learning round and is adaptable for launch at any desired time, such as during the FL training initialization.
Our attack does not cause significant impacts on the training performance and can be launched repeatedly in multiple FL rounds to harvest as many local training samples as possible.

Three factors could significantly impact the performance of our attack---the reconstruction number is restricted by the neuron number of the first linear layer in the latent space; the reconstruction quality is sensitive to the quantity and quality of the auxiliary dataset; and the reconstruction rate relies on the availability of a good statistical estimation of the distribution of LSRs. Fortunately, in practical scenarios, all three factors do not pose a significant barrier for the attacker. Popular machine-learning models commonly feature large linear layers; auxiliary datasets can be easily collected from various online resources and public datasets; and data representations in latent space often follow a Gaussian distribution and are easy to estimate, contributing to better attack performance.

We conducted extensive experiments to evaluate the performance of the proposed attack on the Fashion MNIST (FMNIST) \cite{xiao2017fashion}, CIFAR-10, TinyImageNet \cite{le2015tiny}, and skin cancer dataset (HMNIST) \cite{codella2019skin}. A thorough comparison was made between our attack and existing MIAs and the results show a significant improvement in terms of attack accuracy, reconstruction fidelity, and efficiency when using \sysname. We also evaluated \sysname's performance under different data settings, including variations in data amount and whether the data was iid or non-iid. The results consistently highlighted the success of \sysname across all these settings. Particularly noteworthy is the results demonstrate that the attacker can effectively launch a targeted attack using his collected dataset over a specific class. 

\subsection{Contributions}

This paper makes the following contributions:

\begin{enumerate}

\item  We identify the latent space of a machine learning model as the pivotal layer for launching an MIA to breach user data privacy in federated learning systems. This insight motivates us to focus on the privacy risks posed by individual components within the model architecture, enabling us to devise a more efficient and effective attack strategy from a white-hat attacker's perspective.

\item We propose \sysname, a novel MIA launched by a malicious parameter server. \sysname efficiently and accurately reconstructs a large batch of user data samples from the aggregated model updates, given that the federated learning system is under the protection of a robust secure aggregation protocol. 

\item Compared to existing attacks, \sysname demonstrates significantly improved efficiency and scalability as it removes the requirement for expensive per-batch search-based optimization. Moreover, \sysname is stealthier in its approach, as it does not require any modifications to the global model architecture and can be accomplished within one FL training round.

\item  We provide a comprehensive analysis and evaluate the key factors that significantly impact the performance of \sysname. Alongside our analysis, we present several practical attack scenarios of \sysname to promote the need for novel defenses against such advanced attacks. 

\item  We conducted extensive experiments to evaluate the performance of \sysname under diverse settings. We examined \sysname's performance on popular model architectures including Alexnet, VGGnet, ResNet, and ViT, as well as various datasets. The results show the effectiveness, efficiency, and scalability of our attack.

\end{enumerate}

In Table \ref{tab:definition}, we summarize the definitions and notations used in our paper.

\section{Background and Related Work}\label{Related Work}

\subsection{Federated Learning}

Federated learning (FL) allows a set of clients $\mathcal{C}=\{c_1,c_2,\cdots,c_n\}$ to train a global model $G= f_{\theta}: \mathcal{X \to Y}$ on a global dataset $\mathcal{D}=\cup_{i=1}^{n}D_i$ that is distributed along the users where each client $c_i$ holds a local dataset $D_i$ without the need to share these data samples. 
FL is conducted iteratively in rounds until the model parameter $\theta$ converges. In each round $t$, the parameter server $S$ first publishes the global model parameter $\theta^{t}$ to a subset of selected clients $\mathcal{C}^{t}\subseteq \mathcal{C}$. 
Then these clients compute the gradients with their local batches $D_i^t$ as $g_i^t=\frac{1}{\left| D_i^t\right|}\nabla L(\theta_t, D_i^t)$, where $L()$ refers to the loss function.
The clients send their computed updates back to $S$ and the latter will aggregate the updates with the FedSGD algorithm \cite{mcmahan2017communication}:
\begin{equation}
    \begin{aligned}
        \theta^{t+1}=\theta^t-\eta  \sum_{ i:c_i\in\mathcal{C}^t}\frac{\alpha_i}{\left| D_i^t\right|}\nabla L(\theta_t, D_i^t)
    \end{aligned}
\end{equation}
where $\eta$ is the global learning rate and $\alpha_i$ is the weight assigned to client $c_i$. The summation of all weights $\{\alpha_i\}_{i:c_i\in\mathcal{C}^t}$ is 1 and can be adjusted according to the size of local datasets $D_i^t$ to avoid training bias. The clients can also train the received global model $G_t$ for $L^t_i$ local rounds before providing the model updates $\delta^t_i$ to the server. In this case, the server employs the FedAVG algorithm to conduct the training process:
\begin{equation}
    \begin{aligned}
        \theta^{t+1}=\sum_{ i:c_i\in\mathcal{C}^t}\alpha_i \delta^t_i
    \end{aligned}
\end{equation}
In the following sections, we will omit the notation $t$ because our attack is a single-round attack and can be launched in any FL training round.

\begin{table}[t]
 \centering
    \caption{Definition and notations.}
    \small
    \centering
    \begin{tabular}{p{1cm} p{6cm}}
		\toprule
		Symbol & Definition \\
		\midrule
        $c_i$ & Federated learning clients \\
        $n$ & Number of federated learning clients \\
        $G$ & Global model \\
        $\theta$ & Global model parameters\\
        $t$ & Training round \\
        $m$ & Input batch size \\
        $D_i$ & Local datasets \\
        $g_i$ & Individual gradients \\
        $\delta_i$ & Individual model updates \\
        $u_i$ & Masked model updates \\
        SA & Secure Aggregation \\
        $x_i$ & Input samples \\
        $\hat{x_i}$ & Reconstructed samples \\
        $L$ & Loss function \\
        $W^{[2]}$ & Two-layer linear leakage module\\
        $D_{Adv}$ & Auxiliary dataset \\
        $G_{Adv}$ & Adversarial model \\
        MLP & Multi-layer perception \\
        $\hat{Enc}$ & Surrogate encoder \\
        $\hat{Dec}$ & Generative decoder \\
        $\hat{G}$ & Modified global model \\
        LSR & Latent Space Representation \\
        CDF & Cumulative Density Function \\
		\bottomrule
	\end{tabular}
        \vspace{-0.1in}
	\label{tab:definition}
\end{table}

\subsection{Gradient Inversion}

The gradient inversion problem is to find a function that can reverse the individual gradient $g_i$ uploaded by a client $c_i$ back to the local dataset $D_i$ under the FL setting, which can be defined as $D_i\stackrel{?}{=} Reverse(g_i, G)$.

\begin{table*}[t]
    \centering
    \caption{A comparison between different federated learning model inversion attacks.}
    \small
    \centering
    \begin{tabular}{lcccccc}
        \toprule
           & Break Secure & Attacker's & Attack & Attack & Need Auxiliary & Model \\
        Attack & Aggregation? & Capability & Overhead & Scale & Dataset? & Agnostic? \\
        \midrule
        DLG \cite{zhu2019deep}, iDLG \cite{zhao2020idlg} & No & Weak (Curious) & Large & Single image & No & Yes \\
         Inverting Grad \cite{geiping2020inverting} & No & Weak (Curious) & Large & 8 & No & Yes \\
       GradInversion \cite{yin2021see} & No & Weak (Curious) & Large & 48 & No & No (ResNet) \\
       GradViT \cite{hatamizadeh2022gradvit} & No & Weak (Curious) & Large & 8 & No & No (ViT) \\
        APRIL-Optim \cite{lu2022april} & No & Weak (Curious) & Large & Single-image & No & No (ViT) \\
        APRIL-Analytic \cite{lu2022april} & No & Weak (Curious) & Small & Single-image & No & No (ViT) \\
        R-GAP \cite{zhu2020r} & No & Weak (Curious) & Small & Single-image & No & Yes \\
        Leak in FA \cite{dimitrov2022data} & No & Weak (Curious) & Small & 50 & No & Yes \\
         Fishing for data \cite{wen2022fishing} & Yes & Medium (Modify params) & Large & 256 & Yes & Yes \\
        Eluding SecureAgg \cite{pasquini2022eluding} & Yes & Medium (Modify params) & Large & 512 & Yes & Yes \\
        Robbing the fed \cite{fowl2021robbing} & Yes & Strong (Change architect) & Small & 1024+ & Yes & Yes \\
        LOKI \cite{zhao2024loki} & Yes & Strong (Change architect) & Small & 1024+ & Yes & Yes \\
        \midrule
        \textbf{\sysname} & Yes & Medium (Modify params) & Small & 1024+ & Yes & Yes \\
        \bottomrule
    \end{tabular}
    \vspace{-0.1in}
    \label{tab:literature-review}
\end{table*}

\textbf{Optimization-based Gradient Inversion:}
Recent research shows that a \textit{honest-but-curious} attacker can solve the gradient inversion problem by solving the following optimization problem:
\begin{equation}
\begin{aligned}
\arg\min_{\hat{D_i}}[d(\nabla \hat{D_i}-\nabla D_i)+r(\hat{D_i})]
\end{aligned}
\end{equation}
where $\hat{D_i}$ refers to randomly initialized dummy samples, $d()$ refers to the distance function, and $r()$ refers to the regulation function. 
Zhu et al. \cite{zhu2019deep} first identify this problem, and propose the deep leakage from gradient (DLG) attack which chooses the second norm as the distance function, and uses the L-BFGS optimizer \cite{byrd2016stochastic} to solve the optimization problem. 
Then Zhao et al. \cite{zhao2020idlg} improve this attack by proposing an analytical method to recover ground-truth labels from the gradients that help DLG achieve better performance. 
Geiping et al. \cite{geiping2020inverting} further improve the optimization tool and achieve better image reconstruction fidelity, but requires a strong assumption as the labels of the inputs must be known. Yin et al. \cite{yin2021see} focus on reconstructing batched inputs on the ImageNet dataset and ResNet model architecture, making the attack more practical. Hatamizadeh et al. \cite{hatamizadeh2022gradvit} customize the attack for the vision transformer and achieve better performance than previous attacks. Dimitrov et al. \cite{dimitrov2022data} devise the attack applicable to the FedAVG setting, making it deployable on real systems.

However, these optimization-based gradient inversion attacks are computationally costly and need hundreds of optimization iterations to reconstruct one input batch \cite{huang2021evaluating}. Many of them (e.g., \cite{zhu2019deep, zhao2020idlg, geiping2020inverting, hatamizadeh2022gradvit, dimitrov2022data}) only perform well for a single or small batch of images (typically smaller than 16), representing a scalability challenge.

\textbf{Closed-form Gradient Inversion:} Instead of using the costly optimization approach, some other works seek to solve the gradient inversion problem with closed-form derivation.  
Aono et al. \cite{aono2017privacy} analyze the privacy leakage of linear models and show that the inputs to linear layers can be perfectly reconstructed from its gradients, which is referred to as the ``linear leakage". This primitive is later revised and used as a component of many advanced attacks \cite{fowl2021robbing, wen2022fishing, zhao2024loki} including our work. Zhu et al. \cite{zhu2020r} propose an analytical recursive attack on privacy (R-GAP) to reconstruct the input layer by layer back from the output layer, particularly targeting linear and convolutional neural networks (CNNs). Lu et al. \cite{lu2022april} propose an analytical reconstruction attack specialized for the vision transformer (ViT) and can reconstruct high-fidelity images. Unfortunately, all of these analytical attacks are designed for specific model architectures and cannot be generalized for others. Furthermore, they are designed for single gradient inversion and cannot reconstruct large input batches.

\subsection{Secure Aggregation}

Previous gradient inversion attacks rely on the assumption that the attacker (i.e., the parameter server) can obtain the individual gradients $\{g_i\}$ from clients, especially for cross-device federated learning.
As a countermeasure, Bonawitz et al. \cite{bonawitz2017practical} propose the secure aggregation (SA) protocol that masks the original model updates from clients. Specifically, SA is a multi-party computation (MPC) protocol that masks the original model updates $\delta_i$ with random bits from secret sharing but keeps the summation of masked updates $\sum_{i=1}^{n}u_i$ equals to $\sum_{i=1}^{n}\delta_i$. Mathematically, this can be defined as:
\begin{equation}
    \begin{aligned}
        f^{SA}(\delta_1,\delta_2,\cdots, \delta_n)&=(u_1,u_2,\cdots,u_n)\\
        s.t. \sum_{i=1}^{n}u_i&=\sum_{i=1}^{n}\delta_i
    \end{aligned}
\end{equation}
where $f^{SA}$ refers to the abstract function of the SA protocol and the attacker cannot distinguish $u_i$ from a random number. This implies that nothing more than the final aggregated result is leaked to the attacker. 
Because the final result is aggregated from all training samples submitted by all participants, the previous gradient inversion attacks cannot reconstruct meaningful information from such large input batches. 
The SA protocol is also communication efficiency and drop-out resilience, making it one of the most robust defense mechanisms against federated learning privacy inference attacks. 
The follow-up works further improve the performance of the original SA protocol by reducing the communication and computation overheads \cite{bell2020secure, choi2020communication, guo2020v, kadhe2020fastsecagg}, enabling verifiable aggregation \cite{xu2019verifynet}, and bolstering the robustness of the secure aggregation against malicious attacks \cite{pillutla2022robust, burkhalter2021rofl, rathee2023elsa, bell2023acorn, ma2023flamingo}.

\subsection{Breaking the SA}

However, when assuming the parameter server is malicious and capable of modifying the global model $G$ ’s parameters, the SA is breakable. Two general types of attacks have been identified, including the \textit{gradient disaggregation attacks}, aiming to overturn SA's main function by inferring individual model updates $\delta_i$ from the aggregated result $\sum_{i=1}^{n}\delta_i$ with crafted model $\hat{G}$, i.e., $\delta_i=Infer(\sum_{i=1}^{n}\delta_i, \hat{G})$; and the \textit{large batch reconstruction attack} that aims to directly reconstruct the global batch $\cup_{i=1}^{n}D_i$ from the aggregated results $\sum_{i=1}^{n}\delta_i$ with the help of additional adversarial module $M_{adv}$ placing in front of global model $G$ and a representative auxiliary dataset $D_{aux}$, i.e., $\cup_{i=1}^{n}D_i=Reverse(\sum_{i=1}^{n}\delta_i, G\oplus M_{adv}, D_{aux})$.

\textbf{Gradient Disaggregation Attacks:} 
Wen et al. \cite{wen2022fishing} propose a ``fishing strategy'' that magnifies the gradient of a targeted class to dominate the aggregated result with crafted model parameters. The attack generates a close enough approximation of the target gradient out of the final aggregated gradient, which is enough for the attacker to reconstruct input samples through existing optimization-based gradient inversion methods. 
Pasquini et al. \cite{pasquini2022eluding} propose a gradient suppression attack that zeros out all the gradient updates except the target victim's, making the final aggregated result identical to the target's gradient. The attack achieves this by crafting the parameters of a single linear layer and keeping the outputs of that specific layer always smaller than zero, which further leads to zero gradients if the ReLU activation function is used. 
The key limitation of this type of attack is they still use the existing optimization-based gradient inversion methods as their attack component, resulting in poor scalability performance and large computation costs.

\textbf{Large Batch Reconstruction Attacks:} Directly reconstructing the whole input batch from the aggregated result is a challenging task and the existing attacks \cite{fowl2021robbing, zhao2024loki} usually have strong assumptions to accomplish it, including allowing the attacker to modify the pre-agreed model architecture and possessing an auxiliary dataset that has a similar distribution as the training dataset. 
Fowl et al. \cite{fowl2021robbing} modify the global model architecture by attaching an adversarial two-linear-layer module in the front. The attacker can leverage the ``linear leakage" primitive to perfectly reconstruct the original inputs with large batch sizes by customizing the parameters of the adversarial module, which are generated from the statistics of the auxiliary dataset. Zhao et al. \cite{zhao2024loki} improve this attack by changing the linear module to a customized convolutional module. 
As a result, the attacker can recover large batches under a more practical FedAVG setting and can help to identify the belongings of the reconstructed samples. 
The key drawback of these attacks is they require the attacker to change the pre-agreed model architecture, which can be easily detected when the clients employ some integrity-checking mechanisms.

We summarize the pros and cons of existing attacks in Table \ref{tab:literature-review}. We compare the existing MIAs with our proposed attack concerning the attack assumption, overhead, and performance. Notably, our proposed \sysname scales significantly better than the existing optimization-based gradient inversion attacks \cite{zhu2019deep, zhao2020idlg, geiping2020inverting, yin2021see, hatamizadeh2022gradvit, lu2022april} along with a small attack overhead, being model-agnostic, and the ability to launch a targeted attack against a certain class. Compared to the gradient disaggregation attacks \cite{wen2022fishing, pasquini2022eluding}, \sysname does not involve any per-batch optimization process and thus reduces the attack overhead (\sysname can be used to replace the costly optimization-based inversion process after the individual gradient is obtained). Compared to the existing large-batch reconstruction attacks \cite{fowl2021robbing, zhao2024loki}, \sysname assumes weaker attacker capability and is more stealthy and harder to detect.

\section{Threat Model}

In this section, we formalize the attacker's capability and goal. We assume the parameter server is malicious and knows the global model $G$ and its parameters $\theta$ of every round.
We consider the state-of-the-art SA protocol (as used in \cite{bonawitz2017practical,bell2020secure}) is in place and the attacker only sees the already masked model updates $\{u_i\}_{i=1}^n$ from the clients rather than the original ones $\{\delta_i\}_{i=1}^n$. 
The attacker cannot distinguish $u_i$ from a random number but he can obtain the aggregated model update $\sum_{i=1}^{n}\delta_i$ through summing the masked inputs $\sum_{i=1}^{n}\delta_i=\sum_{i=1}^{n}u_i$. We assume the communications between the parameter server and clients are secure and no third party can alter the transmitted messages between them. 
We assume the attacker is able to modify the global model $G$'s parameters but \textbf{not architecture} and knows global training configurations such as the learning rate $\eta$ and weights $\alpha_i$, as in \cite{pasquini2022eluding} and \cite{wen2022fishing}. We also assume the attacker possesses an auxiliary dataset $D_{Adv}$ as a subset of all the training data $D_{Train}$ following the same assumptions in \cite{fowl2021robbing,zhao2024loki}. In practice, the attacker can collect this $D_{Adv}$ by using the existing public resources, manually collecting samples, or even colluding with a fraction of clients.

\begin{figure}[t]
    \centering
    \includegraphics[width=0.42\textwidth]{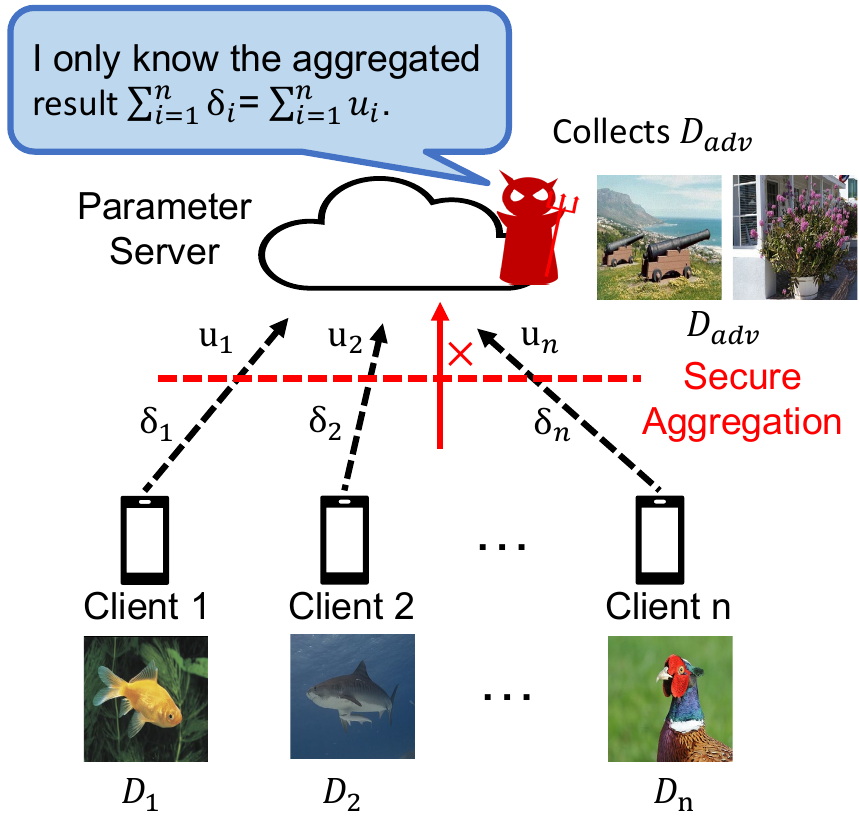}
    \caption{\sysname threat model.}
    \vspace{-0.15in}
    \label{fig: Threat Model} 
\end{figure}

The attack aims to achieve the same goal as \cite{fowl2021robbing,zhao2024loki}, which is to recover the whole global input batch $\cup_{i=1}^{n}D_i$ efficiently and precisely, i.e. $\cup_{i=1}^{n}D_i=Reverse(\sum_{i=1}^{n}\delta_i, \hat{G}, D_{Adv})$. We illustrate our threat model in Figure \ref{fig: Threat Model}.

\section{Attack Preliminaries}\label{Preliminaries}

\subsection{Autoencoder}

Autoencoder is an unsupervised learning technique that helps to learn an informative and compressed data representation \cite{bank2020autoencoders, pu2016variational, he2022masked}. The autoencoder's model architecture contains an encoder $Enc$ and a decoder $Dec$. It is trained to minimize the difference between the original inputs and the reconstructed outputs, i.e., $\arg\min_{\theta}d(x, Dec(Enc(x)))$, where $\theta$ is the autoencoder's parameter vector and $d()$ refers to the error function such as the mean square error. As a result, a fine-tuned autoencoder can almost perfectly reconstruct its model inputs at the outputs, i.e. $x \approx Dec(Enc(x))$, even for the batched inputs $\cup_{j=1}^m x_j$. More specifically, a well-trained autoencoder consists of an encoder 
that encode the input samples to their latent space representations (LSRs), i.e. $\cup_{j=1}^m LSR_j=Enc(\cup_{j=1}^m x_j)$, and a decoder that decodes the LSRs to samples, i.e. $\cup_{j=1}^m \hat{x_j}=Dec(\cup_{j=1}^m LSR_j)$ and $\hat{x_j}$ satisfies $\hat{x_j}\approx x_j$. 

\begin{figure}[t]
    \centering
    \includegraphics[width=0.47\textwidth]{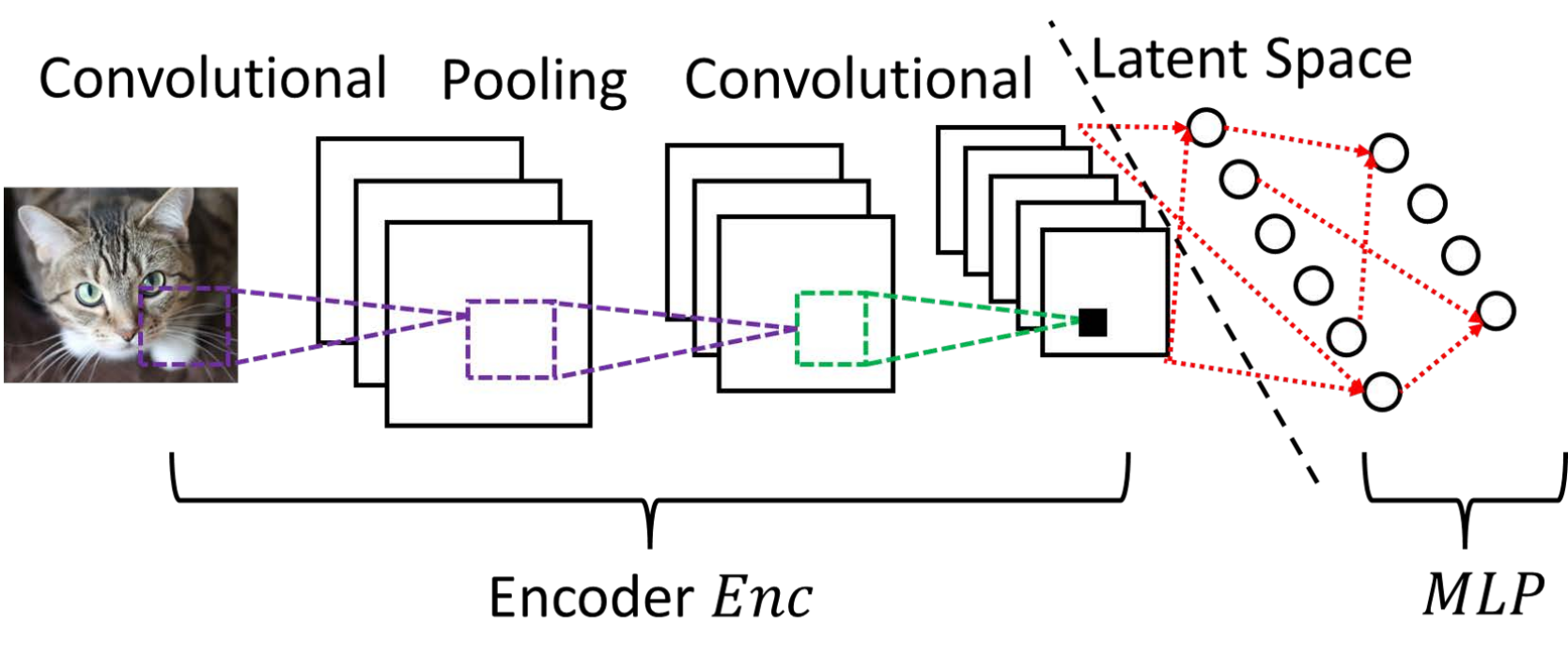}
    \caption{The model architecture of machine-learning classifiers.}
    \vspace{-0.15in}
    \label{fig: Model Architecture} 
\end{figure}

\subsection{Linear Leakage}

Linear leakage refers to a mathematical property that a model with two subsequent linear layers, denoted by $W^{[2]}$,  with a non-linear activation function (e.g. ReLU) in between can be crafted to \textit{perfectly reconstruct} 
 its batched inputs $\cup_{j=1}^m x_j$ from the aggregated gradients of the layers $\sum_{j=1}^m g^{[2]}_j$ with the help of a representative auxiliary dataset $D_{aux}$, i.e. $\cup_{j=1}^m x_j=LinearLeak(\sum_{j=1}^m g^{[2]}_j, W^{[2]}, D_{aux})$ \cite{fowl2021robbing}.

To describe this property, we define the linear module as:
\begin{equation}
    \begin{aligned}
        y&=\delta(w_1 x+b_1)\\
        z&=w_2 y+b_2
    \end{aligned}
\end{equation}
where $x\in \mathbb{R}^d$, $y \in \mathbb{R}^k$, $z \in \mathbb{R}^o$, $w_1 \in \mathbb{R}^{k \times d}$, $b_1 \in \mathbb{R}^k$, $w_2 \in \mathbb{R}^{o \times k}$, $b_2 \in \mathbb{R}^o$, and $\delta$ is the ReLU function. 
Suppose the attacker can accurately estimate the CDF of feature $h(x)=v_h. x$ of the input dataset as $\psi (h(x))$ from the auxiliary dataset $D_{aux}$, where $v_h \in \mathbb{R}^{1\times d}$. Suppose the loss function is $L(x;\theta)$ or simply $L$ where $\theta$ refers to the module parameters. The input batch size is $m$ and can be expressed as $\cup_{i=1}^m x_j=[x_1,x_2, \cdots, x_m]$.

The attacker can craft the linear leakage module in the following steps:
(1) Having the row vectors $w_{1(r)}, r \in \{1,2,\cdots,k\}$ of weight matrix $w_1$ all identical to $v_h$; 
(2) dividing the distribution of the feature $h(x)$ into equally $k$ bins by calculating $h_i=\psi^{-1} (\frac{i}{k})$, which results in having a random variable $h$ the same probability to falls in each bin $[h_s, h_{s+1}]$; (3) assigning the bias vector $b_1$ identical to the opposite values of $h$ vector $[-h_1, -h_2, \cdots, -h_k]$; and (4) letting the row vectors of weight matrix $w_2$ be identical.
After this, the attacker conducts the FL training process to obtain the aggregated gradients and then calculates the following equation to create $k$ bins to reconstruct the input samples, for $r \in \{1,2,\cdots,k\}$:
\begin{equation}
    \label{reconsturction equation}
    \begin{aligned}
        (\nabla_{w_1(r+1)}L-\nabla_{w_1(r)}L)/(\nabla_{b_1(r+1)}L-\nabla_{b_1(r)}L)
    \end{aligned}
\end{equation}
where specially we have $\nabla_{w_1(k+1)}L$ and $\nabla_{b_1(k+1)}L$ equal zero. 

When batch size $m$ is smaller than neuron number $k$, each input sample will only activate and be reconstructed by one bin.
More specifically, sample $x_p$ as the $p^{th}$ smallest one in terms of feature $h(x)$ that falls in the $l^{th}$ bin $[h_l, h_{l+1}]$ alone will be reconstructed by Eq. \ref{reconsturction equation} when $r=l$:
\begin{equation}
\label{recovery equation}
\centering
    \begin{aligned}
        \frac{\nabla_{w_1(l+1)}L-\nabla_{w_1(l)}L}{\nabla_{b_1(l+1)}L-\nabla_{b_1(l)}L}
        &=\frac{\frac{\partial L}{\partial y_{l+1}}\frac{\partial y_{(l+1)}}{\partial w_{1(l+1)}}-\frac{\partial L}{\partial y_{l}}\frac{\partial y_{(l)}}{\partial w_{1(l)}}}{\frac{\partial L}{\partial y_{l+1}}\frac{\partial y_{(l+1)}}{\partial b_{1(l+1)}}-\frac{\partial L}{\partial y_{l}}\frac{\partial y_{(l)}}{\partial b_{1(l)}}}\\
        &=\frac{\sum\limits_{v=1}^{p}\frac{\partial L}{\partial y_{l+1}}x_v-\sum\limits_{v=1}^{p-1}\frac{\partial L}{\partial y_{l}}x_v}{\sum\limits_{v=1}^{p}\frac{\partial L}{\partial y_{l+1}}-\sum\limits_{v=1}^{p-1}\frac{\partial L}{\partial y_{l}}}\\
        &=\frac{\sum\limits_{v=1}^{p}\frac{\partial L}{\partial y_{l}}x_v-\sum\limits_{v=1}^{p-1}\frac{\partial L}{\partial y_{l}}x_v}{\sum\limits_{v=1}^{p}\frac{\partial L}{\partial y_{l}}-\sum\limits_{v=1}^{p-1}\frac{\partial L}{\partial y_{l}}}\\
        &=\frac{\frac{\partial L}{\partial y_{l}}x_p}{\frac{\partial L}{\partial y_{l}}}=x_p
    \end{aligned}
\end{equation}
Note that we have leveraged the property of $\nabla_{w_1(l+1)}L=\frac{\partial L}{\partial y_{l+1}}\frac{\partial y_{(l+1)}}{\partial w_{1(l+1)}}=\sum\limits_{v=1}^{p}\frac{\partial L}{\partial y_{l+1}}x_v$, and $\frac{\partial L}{\partial y_{l+1}}=\frac{\partial L}{\partial y_{l}}$. Their detailed mathematical proof can be found in the Appendix. 

On the other hand, when batch size $m$ is larger than neuron number $k$, linear leakage cannot ensure that one sample is only activated and reconstructed by one bin. In some bins, the samples collide with each other and are mixed together during the reconstruction process, leading to reconstruction failures. Therefore, we regard the neuron number $k$ as the performance bottleneck of the linear leakage primitive. 

\textbf{Linear Leakage in Federated Learning:} In the federated learning system with SA, the parameter server $S$ needs to infer the aggregated gradients $\sum_{j=1}^m g_j$ from the aggregated model updates $\sum_{i=1}^n \delta_i$ in order to launch the linear leakage attack. For the FedSGD system, the two values are identical as $\sum_{i=1}^n \delta_i=\sum_{j=1}^m g_j$. But for the FedAVG system, each model updates $\delta_i$ are trained by clients for several local rounds. Assuming the clients employ the SGD algorithm for local training, the server can only get approximated aggregated gradients $\sum_{j=1}^m \hat{g_j}$ from the aggregated model updates $\sum_{i=1}^n \delta_i$ with analytical tools, resulting in a slightly decreased linear leakage performance, i.e. $\cup_{j=1}^m x_j\approx LinearLeak(\sum_{i=1}^n \delta_i, W^{[2]}, D_{aux})$.

\section{Attack Method}\label{Attack method}

\subsection{Attack Intuition}

Linear leakage provides us with a powerful primitive to reconstruct samples. A straightforward attack strategy is to place a crafted linear leakage module $W^{[2]}$ right in front of the global model $G$ as $G\oplus W^{[2]}$ and publish it to the clients. As a result, when the server receives the aggregated gradients $\sum_{i=1}^n g^{[2]}_i$ from clients, it can reconstruct the input samples with the primitive \cite{fowl2021robbing}. However, as we have discussed in Section \ref{Related Work}, it is too suspicious and can be easily prevented by integrity checking as the attacker needs to change the global model architecture. We abandon this architectural modification approach and examine the \textit{built-in components} of models for potential attack exploitation. 

\begin{figure*}[t]
    \centering
    \includegraphics[width=1\textwidth]{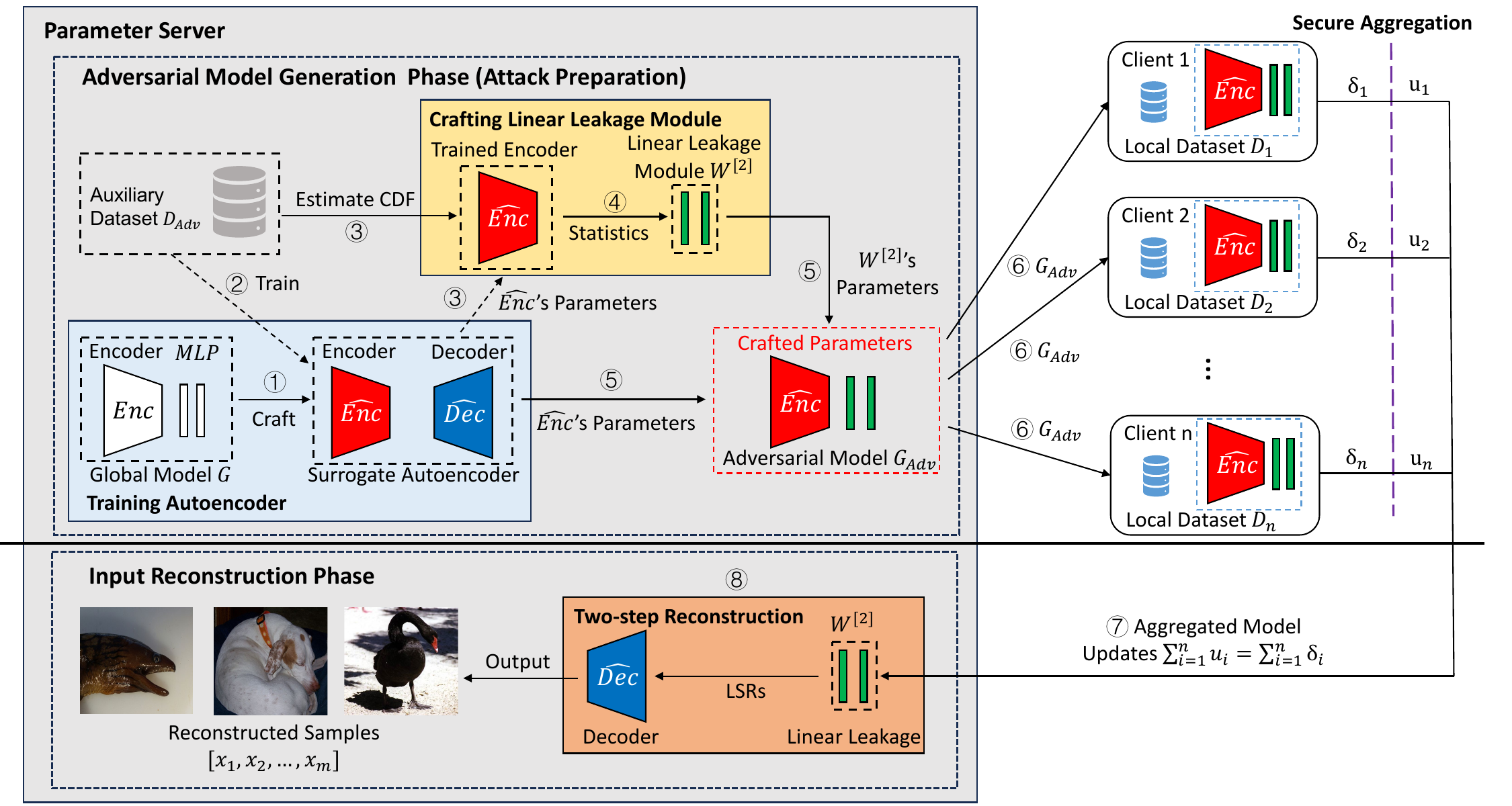}
    \caption{\sysname is a two-phase attack. The first phase is performed locally to produce essential information to conduct the second phase. The second is the actual attack phase for the attacker to interact with the clients and reconstruct their local training samples.}
    \vspace{-0.15in}
    \label{fig: Attackflow} 
\end{figure*}

We observe that machine learning classifiers are commonly composed of a feature extraction encoder $Enc$ followed by a multi-layer perceptron (MLP) in their model architectures, as exemplified in Figure \ref{fig: Model Architecture}. Motivated by this, we target the \textit{latent space} as the key layer to launch the attack based on the following reasons.

\begin{enumerate}
    \item LSRs contain enough information to reconstruct the inputs and are widely regarded as the ``information bottleneck'' within the whole model architecture.
    \item LSRs have relatively lower dimensions and can be processed more efficiently.
    \item In most machine learning models, LSRs are followed up by MLPs, which can be exploited to launch the analytical linear leakage primitive.
\end{enumerate}

\paragraph{Problem Decomposition} Based on our previous findings, we decompose the original complex reconstruction task $\cup_{i=1}^{n}D_i=Reverse(\sum_{i=1}^{n}\delta_i, \hat{G}, D_{Adv})$ into two sub-problems, including firstly reconstruct the LSRs with the linear leakage primitive from the crafted MLPs (in terms of parameters) in the latent space, i.e. $\cup_{j=1}^{m}LSR_j=Reverse(\sum_{i=1}^{n}\delta_i, \hat{W^{[2]}}, D_{Adv})$, then reconstruct the training samples by feeding these LSRs into a fine-tuned decoder, i.e. $\cup_{j=1}^{m}x_j=Dec(\cup_{j=1}^{m}LSR_j)$, where $\cup_{j=1}^{m}x_j$ is identical to $\cup_{i=1}^{n}D_i$. Both two steps only involve matrix computation and feed-forward neural network computations, making the attack super efficient.

\subsection{Attack Overview}

We illustrate the attack flow of \sysname in Figure \ref{fig: Attackflow}. 
\sysname consists of two main phases: the adversarial model generation phase (attack preparation, Steps \textcircled{1}-\textcircled{5}) and the input reconstruction phase (Steps \textcircled{6}-\textcircled{8}). The first phase is conducted locally on the parameter server $S$ and its purpose is to generate an adversarial global model $G_{adv}$ with crafted parameters, as well as a highly accurate generative decoder $\hat{Dec}$. This phase contains two sub-functions including training a crafted surrogate autoencoder $\hat{A}$ and crafting a linear leakage module $W^{[2]}$.
$\hat{A}$ is trained with the auxiliary dataset $D_{Adv}$ and has the same encoder architecture $\hat{Enc}$ as the global model $G$'s encoder $Enc$, as well as a customized decoder $\hat{Dec}$ that has the capability to reconstruct the inputs. 
$W^{[2]}$ is crafted on the MLP layers of the global model $G$, whose necessary parameters and statistics are produced through feeding the auxiliary dataset $D_{Adv}$ to the already-trained encoder $\hat{Enc}$. Finally, the adversarial model $G_{Adv}$ is assembled with the surrogate encoder $\hat{Enc}$ and the linear leakage module $W^{[2]}$. By doing so the adversarial model $G_{Adv}$ has the same architecture as the original global model $G$ but with the necessary parameters to launch our reconstruction attack. This phase is conducted completely offline and covers all the required training efforts.

\begin{figure*}[t]
    \centering
    \includegraphics[width=1\textwidth]{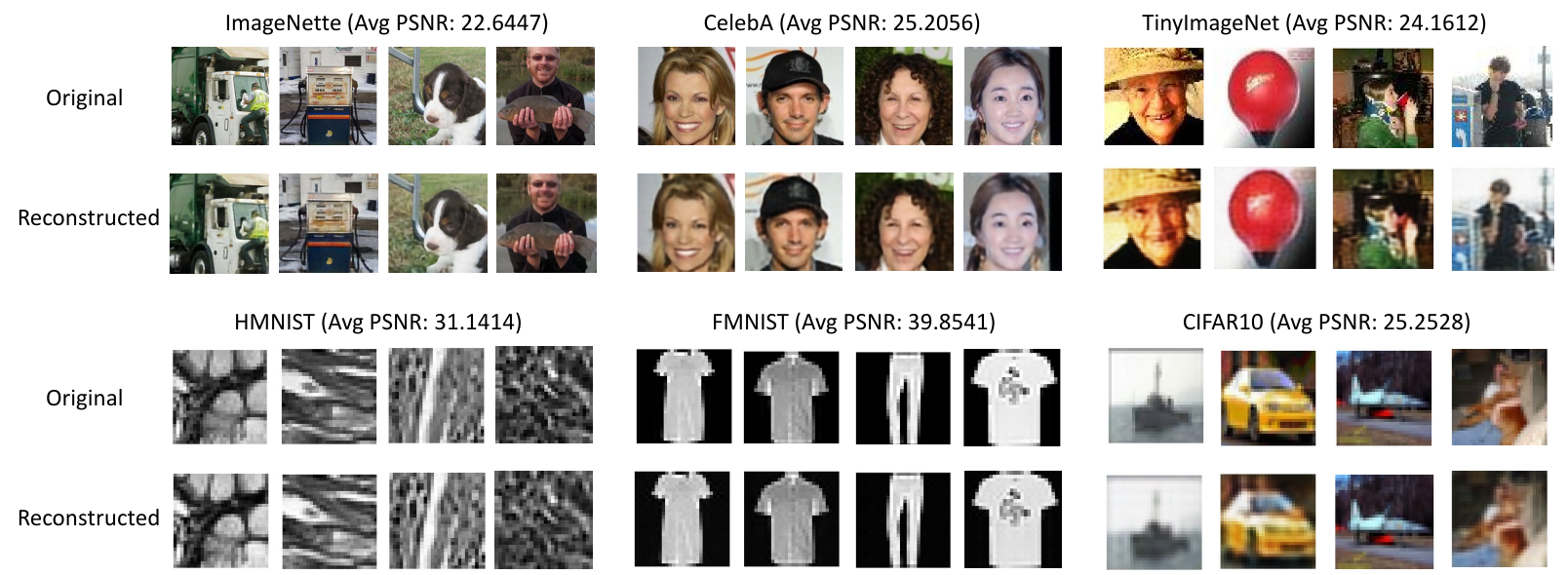}
    \caption{Reconstruction examples. These examples are taken from large reconstruction batches. Full reconstructed batches and more discussions can be found in the Appendix.}
    \vspace{-0.15in}
    \label{fig: Examples} 
\end{figure*}

For the input reconstruction phase, the attacker first distributes the adversarial model $G_{Adv}$ to the clients and awaits their feedback. After receiving the feedback, the attacker examines the model updates of the MLP layers in the adversarial global model to first reconstruct the batched latent space representations (LSRs) through the linear leakage primitive, and then reconstruct the input samples by feeding the LSRs to the trained decoder $\hat{Dec}$. All the operations involved in this actual attack phase only involve linear-complexity calculations and the reconstruction is significantly accelerated.

\subsection{Detailed Workflow}

\textbf{Adversarial Model Generation:} We assume that the global model $G$ has an architecture consisting of an encoder $Enc$ followed by a multi-layer perception (MLP). This assumption is practical and commonly observed in popular image classification models such as CNN-based AlexNet, VGGNet, ResNet, and Vision Transformers.
In step \textcircled{1}, the attacker crafts a surrogate autoencoder $\hat{A}$ consisting of an encoder $\hat{Enc}$ and a decoder $\hat{Dec}$. The encoder $\hat{Enc}$ has the same model architecture as the global model's encoder $Enc$ and the decoder $\hat{Dec}$ is constructed according to the model architecture of $\hat{Enc}$. For example, for an encoder with several convolutional layers, the decoder can have several de-convolutional layers to reconstruct the input. In step \textcircled{2}, the attacker trains the surrogate autoencoder $\hat{A}$ using the auxiliary dataset $D_{Adv}$ with the objective of minimizing the distance between its inputs and outputs. $D_{Adv}$ can be a publicly available dataset or a custom-collected dataset targeting a specific purpose or victim. In step \textcircled{3}, the attacker feeds the auxiliary dataset $D_{Adv}$ to the already trained encoder $\hat{Enc}$ to get the LSRs ($LSR_{Adv}$) of the dataset. The attacker estimates the CDF of the \textbf{\textit{brightness}} (the average value among all pixels) of the $LSR_{Adv}$ and calculates the corresponding bias vector $H=[-h_1, -h_2, \cdots, -h_k]$ of the $k$ bins according to the ``linear leakage'' primitive we described. Here $k$ is identical to the neuron number of the first MLP layer of the global model $G$. 
Then, in step \textcircled{4}, the attacker crafts a linear leakage module $W^{[2]}$ on the first two MLP layers with respect to $H$. More specifically, the attacker has all the elements of the weight matrix of the first layer $w_1$ identical to $\frac{1}{d}$, where $d$ is the dimension of the LSRs; as well as having the bias vector of the first layer $b_1$ equals $H$ and the row vectors of the second weight matrix $w_2$ identical.
Finally, the attacker generates the adversarial global model $G_{Adv}$ by having the parameters of $G_{Adv}$'s encoder identical to $\hat{Enc}$ and the parameters of the MLP layers identical to $W^{[2]}$. 

\textbf{Input Reconstruction:} After generating the adversarial model $G_{Adv}$, the attacker publishes it in step \textcircled{6} to all clients. Then according to the FL framework, in step \textcircled{7} the clients send back their model updates $\delta_i$. Because we assume the SA is in place, the server receives the aggregated model update $\sum_{i=1}^{n}\delta_i$ instead of individual ones. In step \textcircled{8}, the reconstruction module takes the aggregated model update as the input and first uses the linear leakage primitive to recover a batch of LSRs, i.e. $[LSR_1, LSR_2, \cdots, LSR_m]\approx LinearLeak(\sum_{i=1}^n \delta_i, W^{[2]}, D_{Adv})$, where $m$ is the global batch size identical to the cardinality of $\cup_{i=1}^{n}D_i$. Then these LSRs are taken as the inputs to the trained decoder $\hat{Dec}$ and the attacker finally gets $\cup_{j=1}^m\hat{x_j}=\hat{Dec}(\cup_{j=1}^m LSR_j)=\hat{Dec}(\hat{Enc}(\cup_{j=1}^m x_j))$ as the reconstruction outputs.
According to the mathematical property of the linear leakage primitive and autoencoder, $\hat{x_j}$ and $x_j$ are identical or highly similar.

\begin{table*}[t]
    \centering
    \caption{The comparison between \sysname and the existing MIAs. * and $+$ means that we use different thresholds other than 18. For \cite{geiping2020inverting} we select the threshold to be 14 and for \cite{yin2021see} select the threshold to be 10.}
    \small
    \centering
    \begin{tabular}{lcccccccc}
        \toprule
         Attack  & Metric & 1 & 2 & 4 & 8 & 16 & 64 & 256 \\
        \midrule
         \multirow{3}{*}{DLG \cite{zhu2019deep} /iDLG \cite{zhao2020idlg}} & PSNR (dB) & 33.0844 & -- & --& --& --& --& -- \\
          & Time (s) & 127.1043  & -- & --& --& --& --& --  \\
          & Rate (ratio) & 0.634  & -- & --& --& --& --& --  \\
        \midrule
         \multirow{3}{*}{InvertingGrad \cite{geiping2020inverting}} & PSNR (dB) & 15.8407 & 16.2223 & 15.4679 & 14.8693 & -- & -- & --\\
          & Time (s) & 280.126 & 305.6257 & 477.1157 & 866.8414 & -- & -- & -- \\
          & $Rate*$ (ratio) & 0.1999  & 0.12 & 0.0833 & 0.0417 & --& --& --  \\
        \midrule
        \multirow{3}{*}{GradInversion \cite{yin2021see}} & PSNR (dB) & 13.3059 & 12.7896 & 12.0550 & 11.1410 & 10.1029 & -- & -- \\
          & Time (s) & 324.8003 & 341.6113 & 348.2077 & 377.3184 & 462.2559 & -- & --\\ 
          & $Rate^{+}$ (ratio) & 0.99 & 0.95 & 0.85 & 0.65 & 0.245 & -- & --\\
        \midrule
        \multirow{3}{*}{Robbing the Fed \cite{fowl2021robbing}} & PSNR (dB) & 150.568 & 147.9793 & 142.0058 & 134.6393 & 129.1871 & 112.3125 & 103.9919 \\
         & Time (s) & 0.1042 & 0.1124 & 0.1137 & 0.1141 & 0.1169 & 0.1658 & 0.3192 \\
         & Rate (ratio) & 0.89 & 0.925 & 0.91 & 0.8988 & 0.8981 & 0.8743 & 0.8335 \\
        \midrule
        \multirow{3}{*}{ Loki \cite{zhao2024loki}} & PSNR (dB) & - &  85.8121 &  54.2885 &  43.1088 &  41.1882 &  40.4387 &  38.7883 \\
        & Time (s) & - &  3.4335 &  4.4684 &  5.7417 &  8.6886 &  24.8163 &  99.3890 \\
        & Rate (ratio) & - &  1.0 &  0.875 &  0.7666 &  0.7916 &  0.7344 &  0.7109 \\
        \midrule
        \multirow{3}{*}{\textbf{\sysname}} & PSNR (dB) & 29.4192 & 29.4162 & 29.3292 & 29.2977 & 29.1853 & 28.9188 & 27.2986 \\
         & Time (s) & 0.01951 & 0.02154 & 0.02463 & 0.03060 & 0.04134 & 0.09541 & 0.2214 \\
         & Rate (ratio) & 1.0 & 0.999 & 0.993 & 0.9927 & 0.992 & 0.9438 & 0.8464 \\
        \bottomrule
    \end{tabular}
    \vspace{-0.1in}
    \label{tab:Benchmark}
\end{table*} 

\subsection{Efficacy and Efficiency Analysis} 

\textbf{Performance Bottleneck:}
Three main factors affect the performance of \sysname. The first one is the neuron number of the first linear layer $k$, subject to the linear leakage's constraints. However, the neuron numbers of the first linear layers of popular machine learning models are usually very large, typically in the scale of thousands (e.g. 4096 for ImageNet classifiers), which is enough for the attacker to reconstruct hundreds or even a few thousands of samples simultaneously.
In practice, we observe that the attacker can achieve high reconstruction rates ($\ge 0.7$) when the batch size $m$ is lower than $\frac{k}{2}$, i.e. $m<\frac{k}{2}$, instead of the theoretical threshold $k$ because of imperfect estimations and noise.
We observe more collided samples in different bins and gradually dropped reconstruction rates when $m$ exceeds  $\frac{k}{2}$. But we argue that both the theoretical $k$ and practical $\frac{k}{2}$ thresholds are not hard bounds, meaning that the attacker can still reconstruct a portion of local samples even when these thresholds are exceeded, although may only with relatively low reconstruction rates ($\le 0.3$).  

The second factor is the quality of the auxiliary dataset $D_{Adv}$. \sysname can achieve near-perfect reconstruction performance when $D_{Adv}$ can represent the target training dataset $D_{Train}$ well. In practice, the attacker can leverage various online resources including public-available datasets, image-searching tools, and even generative models such as GAN \cite{goodfellow2014generative,karras2019style,park2018data}, and diffusion models \cite{ho2020denoising,song2020denoising,rombach2022high,dhariwal2021diffusion} to help collect the auxiliary dataset. Note that the auxiliary samples do not necessarily need to be highly similar to the targets, all samples in the same format and class can help to improve the reconstruction performance. On the other hand, \sysname enables the attacker to launch a targeted reconstruction even if the auxiliary dataset is biased, amount-deficient, and even skewed to some extent.

The third factor is whether the attacker can have a good estimation of the required statistics  (i.e. the LSR distribution) to craft the linear leakage module. Fortunately, the LSRs in the latent space usually exhibit Gaussian or Laplace distribution and can be accurately estimated by the attacker. Furthermore, because the surrogate autoencoder's training process is fully controlled by the attacker, it can also use the variational autoencoder (VAE), a variant of the autoencoders to regulate the LSRs to follow Gaussian distribution \cite{kingma2013auto,pu2016variational,cemgil2020autoencoding,kingma2019introduction}. In this way, the LSR distribution can be perfectly estimated.

\textbf{Attack Overhead:} The major overhead imposed by \sysname is in the adversarial model generation phase, or more specifically, the training process of the autoencoder. Except this, the other steps in phase one and the second input reconstruction phase only involve analytical operations and are efficient to perform. \sysname allows this autoencoder training process to be conducted fully offline and the attacker can leverage any computation resource to fulfill this. 
The attacker can also resort to finding publicly available pre-trained autoencoders. \sysname is a single-round attack and once the attack preparation phase is finished, the attacker can iteratively launch attacks for multiple rounds, i.e. ``train once, and attack multiple rounds''. 

\textbf{Binary Reconstruction:} \sysname exhibits a binary reconstruction property, meaning that for one specific sample, the reconstruction performance is either good enough to obtain a highly similar reconstructed sample or completely fails to obtain any meaningful content. This is because the major reason for reconstruction failure is the collisions within the reconstruction bins of the linear leakage primitive. The successful samples fall in one bin alone and can be properly reconstructed. But the failed ones fall in the same bin together and their reconstructed images are also mixed and blurred with each other. This property can be validated by the PSNR (attack performance) distribution of different reconstruction batches, which are shown in Figure \ref{fig: PSNR-dist}, as the results exhibit bi-modal distributions because successful samples compose one modal and failed samples compose the other one. 

\textbf{Targeted Attack:} \sysname allows the attacker to launch targeted reconstruction with biased auxiliary dataset $D_{Adv}$ containing limited classes of samples. For example, an auxiliary dataset full of ``dog'' samples can help the attacker reconstruct samples with the ``dog'' label among all input samples. In our experiment, we find that an auxiliary dataset with a few hundred samples in one class can reconstruct new images in the same class with high accuracy. This is particularly useful, as in many cases, the attacker may have a biased auxiliary dataset and is only curious about certain classes of samples.

\begin{table*}[t]
    \centering
    \caption{The reconstruction performance of \sysname over different batch sizes on FedSGD and FedAVG systems.}
    \small
    \centering
    \begin{tabular}{lc cc cc cc cc cc}
        \toprule
         \multirow{2}{*}{Systems}  & \multirow{2}{*}{Datasets}  & \multicolumn{2}{c}{64} & \multicolumn{2}{c}{128} & \multicolumn{2}{c}{256} & \multicolumn{2}{c}{512} & \multicolumn{2}{c}{1024} \\
         & &Rate&PSNR &Rate&PSNR &Rate&PSNR &Rate&PSNR &Rate&PSNR\\
        \midrule
                  & CIFAR-10 & 0.9328 & 28.6453 & 0.8988 & 28.3626 & 0.8464 & 27.2986 & 0.7126 & 26.7407 & 0.4841 & 25.2302 \\
        FedSGD    & FMNIST & 0.9402 & 39.4908 & 0.9057 & 38.6366 & 0.7882 & 34.2213 & 0.6613 & 33.9857 & 0.4173 & 28.2084 \\
        (iter=1)  & HMNIST & 0.9619 & 28.3522 & 0.9150 & 26.9998 & 0.8134 & 24.6165 & 0.6719 & 23.2551 & 0.4271 & 22.4616 \\
                  & TinyImageNet & 0.8906 & 25.0072 & 0.8804 & 23.2042 & 0.8136 & 22.9310 & 0.6826 & 22.6568 & 0.5252 & 22.3976 \\
                  & ImageNette & 0.8875 & 22.8267 & 0.8132 & 22.7786 & 0.7136 & 22.5842 & 0.5852 & 22.6251 & 0.4155 & 22.4451 \\
                  & CelebA & 0.9234 & 23.3090 & 0.8656 & 23.1878 & 0.7625 & 23.1235 & 0.5871 & 22.6476 &  0.3721 &  22.3869 \\
       \midrule
                & CIFAR-10 & 0.9231 & 28.3490 & 0.8770 & 28.0728 & 0.8006 & 27.4762 & 0.6129 & 25.2569 & 0.4709 & 24.7438 \\
        FedAVG  & FMNIST  & 0.9687 & 39.2831 & 0.9063 & 38.6342 & 0.8068 & 36.7750 & 0.6641 & 32.1269 & 0.4146 & 28.3235 \\
        (iter=3) & HMNIST & 0.9434 & 28.5524 & 0.9111 & 26.9872 & 0.7246 & 24.7298 & 0.6426 & 22.9255 & 0.5073 & 22.0577 \\
                & TinyImageNet & 0.9070 & 23.2623 & 0.8203 & 23.0105 & 0.7488 & 22.8969 & 0.6804 & 22.6052 & 0.5210 & 22.2692 \\
                & ImageNette & 0.8678 & 22.8635 & 0.8016 & 22.7005 & 0.7141 & 22.6861 & 0.5640 & 22.6489 & 0.3541 &  22.1800 \\
                & CelebA & 0.9188 & 23.1996 & 0.8453 & 23.1373 & 0.7796 & 23.2074& 0.5492 & 22.6717 &  0.3529 &  22.3813 \\
        \midrule
             & CIFAR-10 & 0.9121 & 28.0774 & 0.8412 & 27.7808 &  0.8010 & 27.9652 & 0.6401 & 26.1579 & 0.4507 & 24.7552 \\
         FedAVG & FMNIST & 0.9046 & 38.7166 & 0.8984 & 37.8936 & 0.7421 & 34.0984 & 0.6308 & 34.2613 & 0.4355 & 28.8201 \\
         (iter=5) & HMNIST & 0.9551 & 28.2060 & 0.9089 & 27.0392 & 0.7651 & 24.5628 & 0.5439 & 22.1618 & 0.4628 & 22.1073 \\
            & TinyImageNet & 0.8917 & 23.2300 & 0.8125 & 23.1355 & 0.6804 & 22.6053 & 0.5527 & 22.4353 & 0.5261 & 22.4493 \\
            & ImageNette &  0.8615 &  22.7716 &  0.7953 &  22.6870 &  0.7121 &  22.6863 &  0.5883 &  22.6171 &  0.3506 &  22.3832 \\
            & CelebA & 0.9125 & 23.2661 & 0.8559 & 23.1304 & 0.7547 & 23.1276 & 0.5931 & 23.0004 &  0.3524 &  22.4913 \\
       \bottomrule
    \end{tabular}
    \vspace{-0.1in}
    \label{tab:reconstruction-rate-batches-systems}
\end{table*}

\begin{table*}[t]
    \centering
    \caption{The reconstruction performance of \sysname over different models and batch sizes.}
    \small
    \centering
    \begin{tabular}{l cc cc cc cc cc}
        \toprule
         \multirow{2}{*}{Models}  & \multicolumn{2}{c}{64} & \multicolumn{2}{c}{128} & \multicolumn{2}{c}{256} & \multicolumn{2}{c}{512} & \multicolumn{2}{c}{1024} \\
         &Rate&PSNR &Rate&PSNR &Rate&PSNR &Rate&PSNR &Rate&PSNR\\
        \midrule
        \multirow{1}{*}{Alexnet (512)} & 0.9146 & 29.0523 & 0.8383 & 28.8489 & 0.7308 & 28.5446 & 0.4974 & 27.2409 & -- & -- \\
       \multirow{1}{*}{Resnet (512)} & 0.9162 & 27.9498 & 0.8408 & 27.6572 & 0.7136 & 26.7230 & 0.4869 & 24.1208 & -- & -- \\
        \multirow{1}{*}{ViT (512)} & 0.7656 & 21.3008 & 0.6641 & 20.9629 & 0.5391 & 21.0911 & 0.3809 & 20.8525 & --& -- \\
       \multirow{1}{*}{CNN (1024)} & 0.9328 & 28.6453 & 0.8988 & 28.3626 & 0.8464 & 27.2986 & 0.7126 & 26.7407 & 0.4841 & 25.2302 \\
       \multirow{1}{*}{VGG (1024)} & 0.9572 & 28.2689 & 0.9191 & 28.2333 & 0.8463 & 27.7959 & 0.7301 & 26.9002 & 0.5032 & 25.0889 \\
       \bottomrule
    \end{tabular}
    \vspace{-0.1in}
    \label{tab:recover-rate-models-batches}
\end{table*} 
\section{Evaluation}\label{Evaluation}

\subsection{Experiment Settings}

We implemented \sysname on the PyTorch platform. We run all the experiments on a server equipped with an Intel Core i7-8700K CPU\@ 3.70GHz$\times$12, two GeForce RTX 2080 Ti GPUs, and Ubuntu 18.04.3 LTS. 

We considered three important evaluation metrics including the reconstruction batch size, reconstruction rate, and the peak signal-to-noise ratio (PSNR) score. The batch size refers to the global batch size, i.e. the cardinality of the union of all the local datasets $\left|\cup_{i=1}^{n}D_i\right|$. This can be interpreted as the multiplication of local batch sizes and the number of clients $n$. For example, a global batch with 512 samples can be uploaded by 512 mobile clients with each client having one sample, by 16 clients with each having 32 samples, or just by one client with 512 samples. By default, we consider the system to contain 8 clients in one FL training round.
The reconstruction rate is the ratio between the successfully reconstructed samples and the total samples.
The definition of the successfulness of reconstructing a sample is by calculating the PSNR score between the original input sample and the reconstructed one, and checking whether the score exceeds a certain threshold $th$. In our work, we take $th=18$ because this threshold is enough for the attacker to distinguish the meaningful contents from the reconstructed figures clearly. The PSNR score is a widely adopted metric to quantify reconstruction quality for images and video subject to lossy compression. It can be expressed as $PSNR=20 \log_{10}(\frac{\max_I}{\sqrt{MSE}})$, where $\max_{I}$ refers to the maximum image pixel value and $MSE$ refers to the mean square error. In this work, we use it to measure the performance of \sysname, following the convention of the existing papers \cite{geiping2020inverting,fowl2021robbing, wen2022fishing,zhao2023secure}. 

We implemented \sysname on the Fashion MNIST (FMNIST) \cite{xiao2017fashion}, Colorectal Histology MNIST (HMNIST) \cite{codella2019skin}, CIFAR-10 \cite{krizhevsky2009learning}, TinyImageNet \cite{deng2009imagenet}, { CelebA \cite{liu2018large}, and ImageNette \cite{deng2009imagenet} datasets.} 
Their detailed introduction can be found in the Appendix. 
For each dataset, we randomly selected a subset of the training set (containing \{1\%, 3\%, 10\%, and 100\%\} of total samples) as the auxiliary dataset and aimed to reconstruct the whole evaluation set, making sure there is \textbf{no intersection} between them. 
We evaluated \sysname's performance on common machine learning model architectures including the convolutional neural network (CNN), AlexNet \cite{krizhevsky2012imagenet}, VGGNet \cite{simonyan2014very}, ResNet \cite{he2016deep}, and ViT \cite{dosovitskiy2020image}. 
For each result, we repeated our experiment 5 times to eliminate uncertainty and noise. In Figure \ref{fig: Examples}, we demonstrate several reconstructed examples over the four datasets. We plot the original images in the first row and the reconstructed ones in the second row along with the average PSNR scores. More reconstruction figures with larger batch sizes (from Figure \ref{fig: Batch-example-cifar} to \ref{fig: Batch-example-tiny}) can be found in the Appendix.

\begin{figure*}[t]
    \centering
    \subfigure[ FL training performance with and without \sysname attack.]{\includegraphics[width=0.242\textwidth]{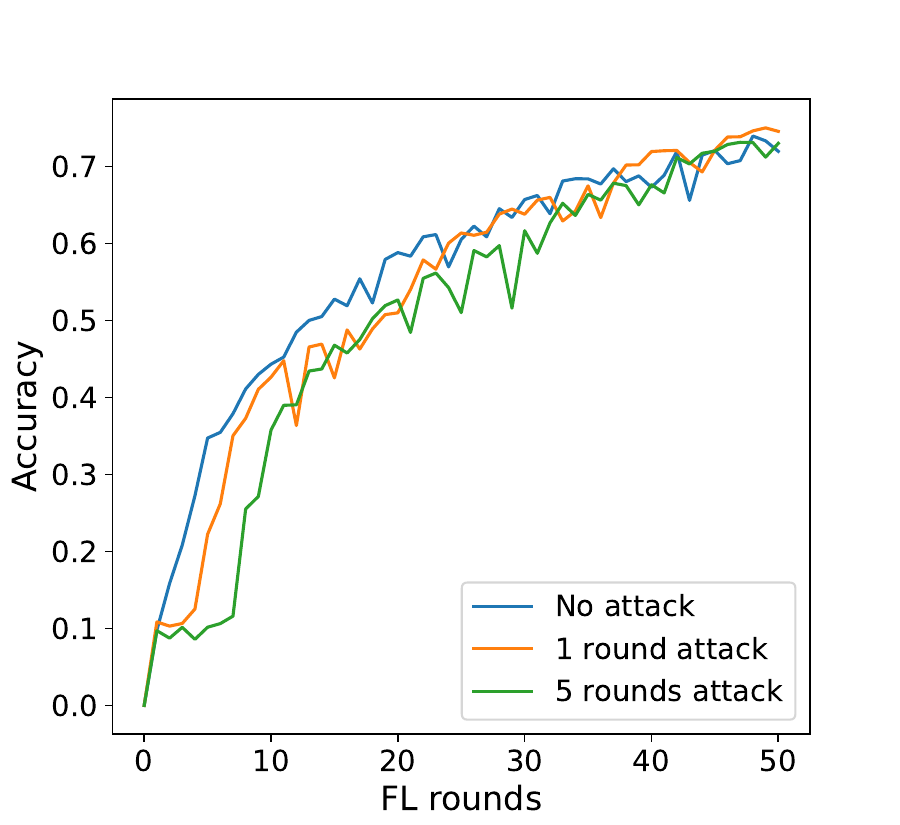}}
    \subfigure[ Reconstruction PSNR scores over different FL client numbers.]{\includegraphics[width=0.242\textwidth]{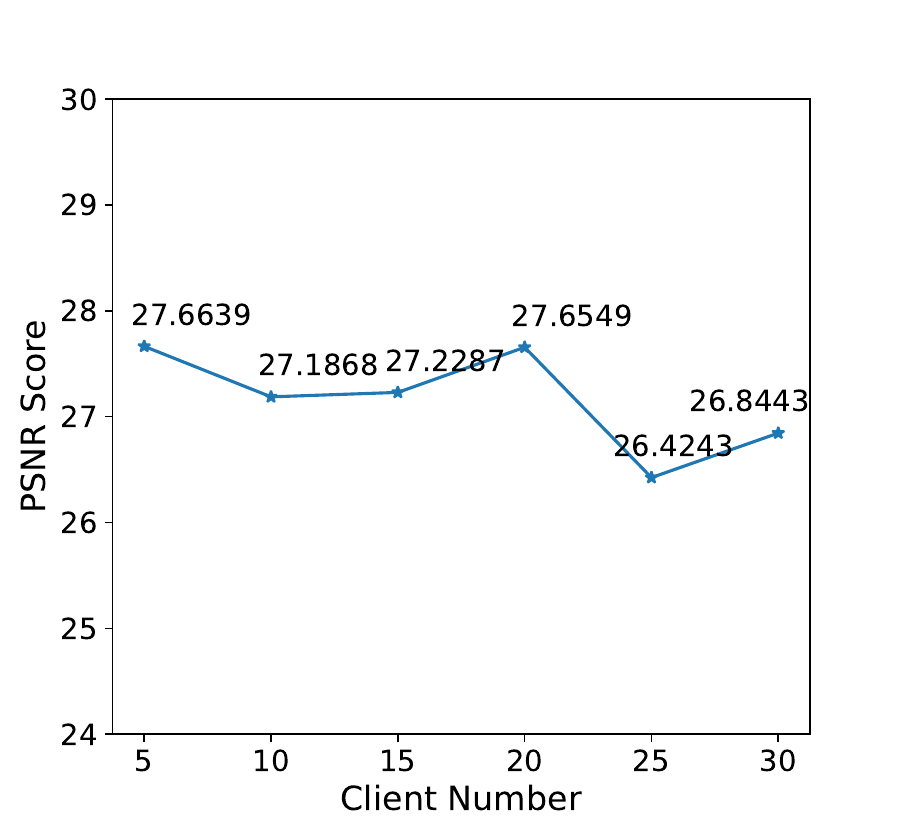}}
    \subfigure[ Reconstruction rates over different FL client numbers.]{\includegraphics[width=0.242\textwidth]{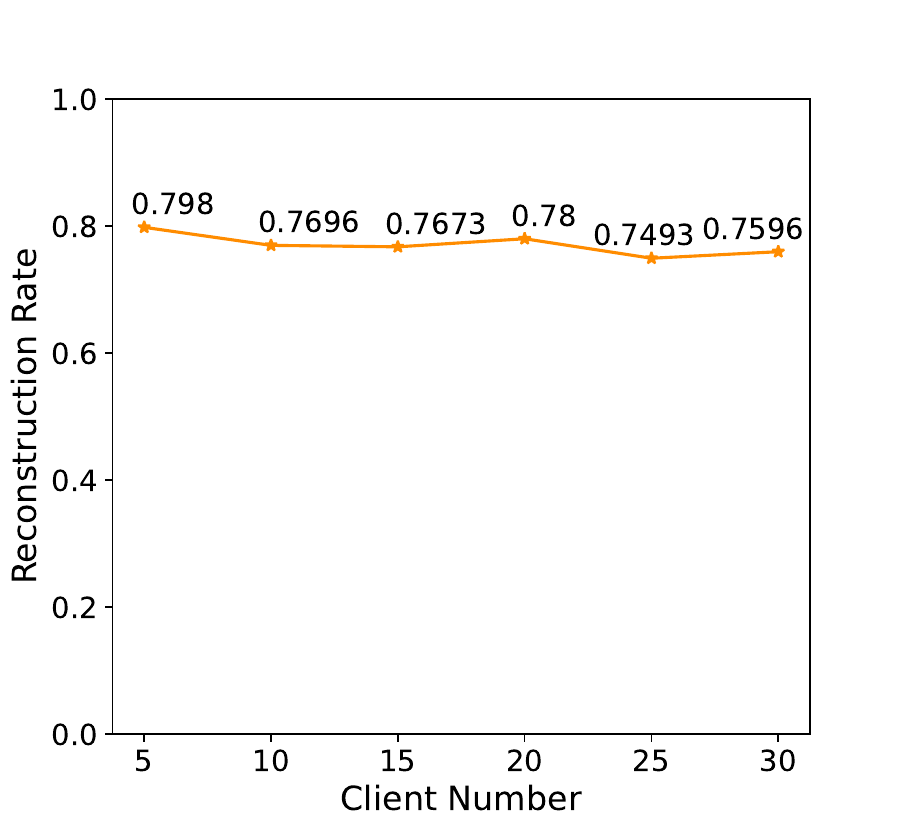}}
    \subfigure[ Attack time (s) over different FL client numbers.]{\includegraphics[width=0.242\textwidth]{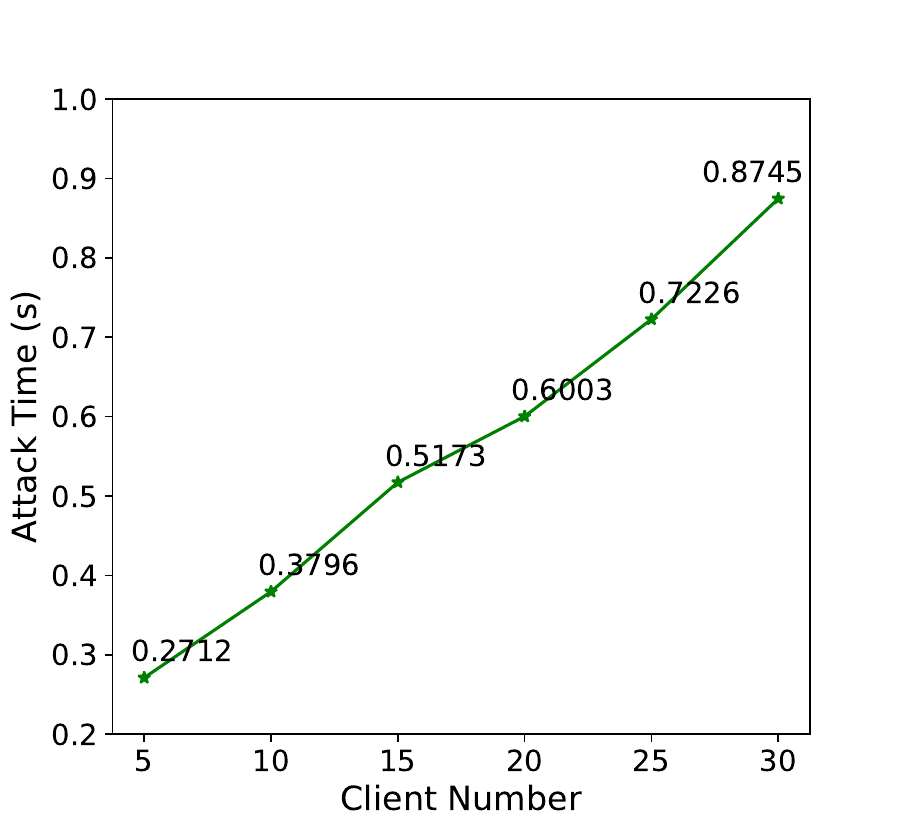}}
    \caption{\sysname's attack performance over different federated learning settings.}
    \label{fig: FL-attack-performance} 
    \vspace{-0.15in}
\end{figure*}

\subsection{Benchmark Comparison}

We implemented and compared \sysname with state-of-the-art 
model inversion attacks that have publicly available artifacts, including the DLG \cite{zhu2019deep}, iDLG \cite{zhao2020idlg}, GradInversion \cite{yin2021see}, Inverting Gradient \cite{geiping2020inverting}, robbing the fed \cite{fowl2021robbing}, { and Loki \cite{zhao2024loki}.} Among them, DLG/iDLG, GradInversion, and Inverting Gradient attacks are well-received optimization-based gradient inversion attacks, which are used as the fundamental building blocks of more advanced gradient disaggregation attacks including fishing for user \cite{wen2022fishing}, and eluding secure aggregation \cite{pasquini2022eluding} attacks. Robbing the Fed \cite{fowl2021robbing} { and Loki \cite{zhao2024loki}} represent the most recent large-batch reconstruction attacks that aim to reconstruct large input batches but at the cost of modifying model architectures.
To make a fair comparison, we re-implemented all these attacks on the CIFAR-10 dataset and with the CNN model architecture. We focused on the reconstruction rate, average PSNR score, and attack time (the time to reconstruct one batch of inputs) metrics to evaluate the effectiveness and efficiency of the attacks on reconstructing sample batches whose sizes ranged from \{1, 2, 4, 8, 16, 64, 256\}. 
We summarize the experiment results in Table \ref{tab:Benchmark}. 

From the experiment results, we observe that \sysname outperforms all other attacks in terms of reconstruction rate and attack time while keeping decent PSNR scores. More specifically, the optimization-based attacks \cite{zhu2019deep, zhao2020idlg,geiping2020inverting,yin2021see} suffer from super long reconstruction time (hundreds of seconds), poor reconstruction rate for even small batch sizes, and low PSNR scores. We also experienced large uncertainty during our implementations of these methods because they rely on search-based optimization methods and we got completely different results with different initialization settings. Note that although the batch sizes we used in this experiment have reached or exceeded the performance upper bound of these optimization-based attacks, they have not yet reached the performance bottleneck of \sysname. In comparison, Robbing the fed attack \cite{fowl2021robbing} achieves good reconstruction rates with large batch sizes, comparable attack efficiency (reconstruct samples in milliseconds), and even better PSNR scores compared to \sysname. This is because Robbing the Fed is a closed-form attack and no optimization process is required. { The other closed-form attack Loki also achieves good reconstruction rates and even better PSNR scores. However, its attack efficiency is significantly worse and requires tens of seconds when the batch size becomes large (as shown in Table \ref{tab:Benchmark}). This is because it adopts a complex model crafting and reversion algorithm.
Moreover, as we have discussed in Section \ref{Related Work}, both Robbing the Fed and Loki attacks adopt a much stronger assumption that requires the attacker to modify the pre-defined model architecture, i.e., adding extra modules before the original model architecture, making them easy to detect. In contrast, \sysname abandons this assumption and still obtains comparable or even better attack performance. }



\subsection{Large Batch Recovery Performance}

We evaluated \sysname's performance with global batch sizes ranging from \{64, 128, 256, 512, 1024\} to validate \sysname's performance on reconstructing large global batches under both the FedSGD and FedAVG settings over different datasets. We changed the number of local training iterations that the clients conducted ranging from \{1,3,5\}. We used a convolutional neural network (CNN) as the target model architecture.
The experiment results are shown in Table \ref{tab:reconstruction-rate-batches-systems}. 

We find that \sysname achieves high reconstruction rates and PSNR scores and there is no obvious performance pitfall when the global batch size is smaller than 512. We observe that both the reconstruction rates and PSNR scores are monotonically decreasing with respect to larger batch sizes, which is consistent with our theoretical results, as larger reconstruction batches increase the probability of reconstruction collisions and failures. We also find that when clients conduct more local training iterations, the attack performance slightly decreases. This is because, with more local iterations, the accuracy of the approximated aggregated gradients from the aggregated model updates decreases. But in general, under all assumptions, \sysname achieves decent attack performance and is not affected by whether the system employs the FedSGD or FedAVG algorithms. 

\begin{table*}[h]
    \centering
    \caption{The reconstruction performance of \sysname over different amounts of data.}
    \small
    \centering
    \begin{tabular}{lcccccccc}
        \toprule
         \multirow{2}{*}{Batch}  & \multicolumn{2}{c}{1\% (500 samples)} & \multicolumn{2}{c}{3\% (1500 samples)} & \multicolumn{2}{c}{10\% (5000 samples)} & \multicolumn{2}{c}{100\% (50000 samples)} \\
         & Rate & PSNR & Rate & PSNR & Rate & PSNR & Rate & PSNR \\
        \midrule
        16 & 0.9808 & 25.1174 & 0.9824 & 25.4396 & 0.9809 & 27.9021 & 0.9827 & 29.4134 \\
         32 & 0.9537 & 24.9192 & 0.9569 & 25.2514 & 0.9550 & 27.7217 & 0.9586 & 29.1968 \\
       64 & 0.9090 & 24.8604 &  0.9136 & 25.1489 & 0.9143 & 27.6032 &  0.9146 & 29.0523 \\
       128 & 0.8243 & 24.6688 & 0.8316 & 24.9584 & 0.8317 & 27.3481 & 0.8313 & 28.8489 \\
       256 & 0.6873 & 24.3534 & 0.6923 & 24.6209 & 0.7018 & 26.8835 &  0.7308 & 27.2409 \\
        \bottomrule
    \end{tabular}
    \vspace{-0.1in}
    \label{tab:recover-rate-data-amount}
\end{table*} 

\begin{table*}[h]
    \centering
    \caption{The reconstruction performance of \sysname over different numbers of classes of data.}
    \small
    \centering
    \begin{tabular}{lcccccccc}
        \toprule
         \multirow{2}{*}{Batch}  & \multicolumn{2}{c}{1 class} & \multicolumn{2}{c}{2 classes} & \multicolumn{2}{c}{3 classes} & \multicolumn{2}{c}{10 classes} \\
         & Rate & PSNR & Rate & PSNR & Rate & PSNR & Rate & PSNR \\
        \midrule
        16 & 0.9818 & 25.1320 & 0.9856 & 24.8238 & 0.9753 & 25.2728 & 0.9827 & 29.4134\\
         32 & 0.9557 & 25.0041 & 0.9701 & 24.8306 & 0.9505 & 25.2147 & 0.9586 & 29.1968\\
       64 & 0.8984 & 24.8337 & 0.9128 & 24.7875 & 0.8971 & 25.1607 & 0.9146 & 29.0523 \\
       128 & 0.8307 & 24.5928 & 0.8568 & 24.5584 & 0.8229 & 25.0155 & 0.8313 & 28.8489\\
       256 & 0.6914 & 24.1795 & 0.7214 & 24.0819 & 0.6615 & 24.5275 & 0.7308 & 27.2409\\
        \bottomrule
    \end{tabular}
    \vspace{-0.1in}
    \label{tab:recover-rate-data-class}
\end{table*} 

\begin{table}[h]
    \centering
    \caption{The reconstruction performance of \sysname over data skew.}
    \small
    \centering
    \begin{tabular}{lccccc}
        \toprule
         Training Data  & \multicolumn{2}{c}{Testing Data 1} & \multicolumn{2}{c}{Testing Data 2}\\
        \midrule
         \includegraphics[width=0.15\linewidth]{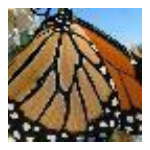}  & \multicolumn{2}{c}{\includegraphics[width=0.15\linewidth]{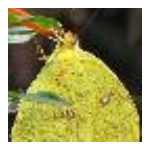}} & \multicolumn{2}{c}{\includegraphics[width=0.15\linewidth]{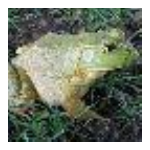}}\\
        \midrule
         \multirow{2}{*}{Batch}  & \multicolumn{2}{c}{Intra-class Skew} & \multicolumn{2}{c}{Inter-class Skew}\\
         & Rate & PSNR & Rate & PSNR \\
        \midrule
         16 &  0.9851 &  23.2206 &  0.7025 &  20.2722 \\
         32 &  0.9750 &  23.1560 &  0.6625 &  20.1722 \\
         64 &  0.9194 &  22.4435 &  0.6468 &  19.7398 \\
         128 &  0.8463 &  22.2957 &  0.6057 &  19.6457 \\
         256 &  0.7929 &  22.1004 &  0.4678 &  19.4764 \\
        \bottomrule
    \end{tabular}
    \vspace{-0.1in}
    \label{tab:recover-rate-data-skew}
\end{table} 

We also evaluated \sysname's performance concerning different model architectures under the FedSGD setting on the CIFAR-10 dataset. The results are shown in Table \ref{tab:recover-rate-models-batches}. We include the neuron number of the first linear layer $k$ beside the model architectures and find that the reconstruction rate is largely affected by it. The models with larger $k$ have better reconstruction rates for a fixed batch size except for a few outliers. We observe that \sysname achieves good reconstruction rates when the batch sizes are smaller than half of the neuron number $k$, in line with our analysis. In general, \sysname achieves decent attack performance on most model architectures, with only achieving a slightly worse attack performance on the ViT, demonstrating that our attack can be applied to different architectures and is model agnostic.

\subsection{Performance over Different FL Settings}

In Figure. \ref{fig: FL-attack-performance}, we demonstrate \sysname's performance under different federated learning settings. We first investigated the overall training accuracy of the FL system with and without our attack. We trained the FL system with 8 clients using the CNN classifier on the CIFAR-10 dataset for 50 rounds. We consider two different attack scenarios, including the attacker only launching a single-round attack in the first training round, and the attacker continuously launching attacks in the first 5 training rounds. From the results, we find that the convergence speed of the FL training process becomes slower in the initial training rounds when the attacks are launched, but the final training accuracy is not affected. In our experiment, the trained model obtained 0.7298 accuracy without any attack, 0.7354 accuracy under a single-round attack, and 0.7301 accuracy under a 5-round attack. 

We also investigated \sysname's performance over different numbers of FL clients. We implemented our attack on the CIFAR-10 dataset and fixed the reconstruction batch size to 300. We increased the FL client number from 5 to 30, which was \{5, 10, 15, 20, 25, 30\}, and evaluated the PSNR scores, reconstruction rates, and elapsed time (in seconds). From the results, we find that the PSNR score and reconstruction rate are slightly affected by the FL client number and remain at a decent high level. This indicates that the reconstruction performance is not affected and the reconstructed images are of good quality. However, we observe that the attack time is almost linear increasing with respect to the FL client number. This is reasonable because more clients involve more communication and computation overhead within the FL system and the complexity of our inversion attack increases accordingly.




\subsection{Data Deficiency and Bias}

The quality of the auxiliary dataset $D_{Adv}$ is an important factor that impacts the performance of \sysname. The ideal situation is that $D_{Adv}$ has the same distribution and can represent the training dataset well. However, in practice, there is usually data deficiency and bias and we evaluate the impacts of them in this section.

\textbf{Data Deficiency}: We varied the amount of data available to the attacker for launching \sysname and evaluated its performance on the CIFAR-10 dataset using the AlexNet model. We considered scenarios where the attacker possesses different proportions of the total training data in the auxiliary dataset, ranging from $1\%$, $3\%$, $10\%$, to $100\%$ (equivalent to 500, 1500, 5000, to 50,000 images). The attack performance was tested on the entire validation set, which consists of 10,000 images. Given that the first linear layer of the AlexNet model has 512 neurons, we varied the reconstruction batch size from 16 to 256 to ensure a reasonable reconstruction performance. 

\begin{table*}[h]
    \centering
    \caption{The reconstruction performance of \sysname when the system is protected by DP.}
    \small
    \centering
    \begin{tabular}{lcccccccc}
        \toprule
         \multirow{2}{*}{Budget/Batch}  & \multicolumn{2}{c}{64} & \multicolumn{2}{c}{128} & \multicolumn{2}{c}{256} & \multicolumn{2}{c}{512} \\
         & Rate & PSNR & Rate & PSNR & Rate & PSNR & Rate & PSNR \\
        \midrule
        No DP & 0.9328 & 28.6453 & 0.8988 & 28.3626 & 0.8464 & 27.2986 & 0.7126 & 26.7407 \\
        $(1, 10^{-5})$ & 0.9253 & 25.2141 & 0.9079 & 25.1279 & 0.8445 & 24.9721 & 0.6855 & 24.6703 \\
        $(1, 10^{-4})$ & 0.9140 & 25.1581 & 0.9019 & 25.1392 & 0.8542 & 25.0477 & 0.6934 & 24.7081 \\
        $(5, 10^{-5})$ & 0.9194 & 25.1359 & 0.9034 & 25.1417 & 0.8359 & 25.1688 & 0.6933 & 24.5526 \\
        \bottomrule
    \end{tabular}
    \vspace{-0.1in}
    \label{tab:dp-attack-performance}
\end{table*} 

We demonstrate the attack performance in table \ref{tab:recover-rate-data-amount}. We observe that the number of available samples has little impact on the reconstruction rate, as it remains relatively stable at high values across the full range of data availability. The PSNR score does decrease slightly when the number of available samples decreases, but it remains in a decent range. Specifically, even if the attacker only has 1\% samples (500 images in total and 50 images for each class) of the total training dataset, \sysname still achieves very high reconstruction rates and PSNR scores in our experiment. This indicates that \sysname is a practical attack and the attacker only needs to collect or generate a few hundred samples to obtain a decent attack performance.

\textbf{Data Bias:} We assumed the auxiliary dataset $D_{Adv}$ to have different numbers of data classes from the CIFAR-10 dataset to evaluate \sysname's performance over biased data. We considered the attacker to have \{1,2,3,10\} classes of data samples and evaluated \sysname's performance on these particular classes. For example, if we had the attacker possess the ``dog" samples, we would only evaluate \sysname's performance on reconstructing the ``dog'' samples in the validation set, ignoring the samples from other classes. We focused on the PSNR score and reconstruction rate for batch sizes ranging from 16 to 256, following the same setting as our previous experiments.

Table \ref{tab:recover-rate-data-class} presents the targeted attack's performance. The results show that data bias has a very limited impact on the reconstruction rate, as it remains stable even when the attacker has only a few classes of samples. The decrease in PSNR score is also not significant, and even the worst value remains at a relatively high level. These results demonstrate that \sysname's attack performance is robust against data bias. The attacker can successfully launch a targeted attack on specific classes with minimal performance degradation.

\textbf{Data Skew:} We considered the auxiliary dataset $D_{Adv}$ to contain skewed data from the target images. We considered two data skew cases including intra-class skew and inter-class skew. We conducted our experiment on the TinyImageNet dataset and evaluated the attack reconstruction rates and PSNR scores. We considered the auxiliary dataset to contain 500 ``monarch butterfly'' images. For the intra-class skew, we assumed the target images were 500 ``sulfur butterfly'' images. For the inter-class skew, we assumed the targets were 500 ``frog'' images.

In Table \ref{tab:recover-rate-data-skew}, we demonstrate the attack performance over different data skew settings. We find that the attack performance slightly decreases (but remains decent) under intra-class skew settings, while the attack performance significantly drops under inter-class skew settings. This is because autoencoders can only reconstruct images similar to training samples by design. The results indicate that \sysname can overcome intra-class skew well, but still faces gaps in dealing with inter-class skew.

\subsection{Differential Privacy}

Differential privacy (DP) \cite{dwork2006differential,dwork2014algorithmic,dwork2008differential} has been widely used to protect the training data's privacy in machine learning systems \cite{abadi2016deep, mcmahan2018learning,blanco2022critical,jayaraman2019evaluating,song2021systematic}. It has shown its effectiveness in protecting client-level and data record-level membership privacy for FL systems \cite{mcmahan2018learning} by preventing the attacker from knowing whether one item (either a client or a record) exists in the system. The fundamental idea of DP is to add artificial noise to the model updates before they are sent to the parameter server.
Though DP can protect the FL systems against membership inference attacks by its definition, it is demonstrated to be less effective against the model inversion attacks \cite{na2022closing}.

In this section, we consider the FL system is protected by both the DP and SA protocols, and we examine \sysname's performance on it to check whether our attack can still break them. 
We assume the clients adopt the DP-SGD algorithm \cite{abadi2016deep} during the local training process with different $(\epsilon, \delta)$ privacy budgets. 
In our experiment, we implement DP with the open-source Python-based DP library named Opacus \cite{yousefpour2021opacus}.
We demonstrate the results in Table \ref{tab:dp-attack-performance}.
From the results, we find that the reconstruction rate is slightly affected by the DP mechanism and remains stable at high levels under different privacy budgets. The PSNR scores decrease slightly when DP is employed but not significantly. This shows that \sysname can still reconstruct samples with high accuracy and good scalability performance even when the DP is in place. However, \sysname cannot link the reconstructed samples back to its clients (membership inference), showing that DP can still preserve a certain level of privacy.

\section{Discussion}
\textbf{Privacy by Shuffling:}
\sysname allows a malicious parameter server to reconstruct the whole input batch accurately from the aggregated model updates, demonstrating a serious privacy vulnerability of the SA protocol and federated learning system. However, under the current design, the attacker cannot infer the belongings of these reconstructed samples and attribute them to certain clients.
This property is known as ``privacy by shuffling'', which prevents the attacker from conducting membership inference and shows that the SA protocol can still preserve a certain level of privacy. To further break this privacy guarantee, \sysname can be used in conjunction with the gradient disaggregation attacks \cite{wen2022fishing, pasquini2022eluding} by launching these attacks in the first step to obtain the individual model updates from the victims and then using \sysname as a replacement of the existing gradient inversion mechanisms to boost the reconstruction performance.

\textbf{Two-Linear-Layer Limitation:}
\sysname requires the model to have two consecutive linear layers to craft the linear leakage module. However, we acknowledge that not all machine learning models necessarily have this component in their architectures although most machine learning classifiers contain it. For example, some Resnet-based and ViT-based models only have one linear layer in the latent space and the attacker must add extra linear layers to launch \sysname. Meanwhile, we observe that some non-linear modules in the feature extraction layers such as the convolutional layers and vision transformers may also leak private information analytically, which can help the attacker bypass the two-linear-layer limitation. We leave this as our future work to investigate.


\textbf{Attack Stealthiness}: 
\sysname is a single-round attack that can be executed at any stage of FL training. To enhance attack stealthiness while maintaining effective performance, the attacker may choose to launch the attack during the initial or the first training rounds. In these rounds, the model parameters can be initialized with arbitrary patterns, making it challenging for defenders to distinguish between adversarial and benign parameters. From a performance perspective, the parameter server receives relatively large aggregated gradients during these initial rounds, as the global model has not yet converged, which facilitates more accurate linear leakage calculations and improves the model inversion performance.

\textbf{Potential Countermeasure}: The data synthesis method could be a potential countermeasure against our novel attack. The fundamental idea is to have each client generate a \textit{mask set} that hides the original sensitive data samples. These mask sets ensure that the mask samples within them rather than the original local samples owned by individual clients are reconstructed during the reconstruction attack.
At the same time, the mask sets shall not affect the federated learning training performance. 
We consider the guided diffusion model a promising tool for generating such mask sets. But we leave the detailed technical design as the future work.

\section{Conclusion}

In this paper, we propose \sysname, a powerful MIA that breaks the strong SA protocol by reconstructing the whole global batch possessed by the clients efficiently from the already masked and aggregated model updates. \sysname launches the inversion attack from a new perspective by delving into the detailed architecture of the global model and decomposing the complex model inversion problem into two steps: an LSR reconstruction step and an input generation step, based on the observation that the latent space is the critical layer to breach the privacy. \sysname uses a closed-form ``linear leakage'' primitive to conduct the first step and a fine-tuned generative decoder for the second, making it highly efficient and suitable for large-scale reconstruction. \sysname is also a very stealthy attack as it does not modify the model architecture and can be conducted in any FL training round. With these distinct features, \sysname represents a potent and inconspicuous approach to breach privacy in FL systems, prompting the need for more robust defense mechanisms against such advanced attacks.


\bibliographystyle{ieeetr}
\bibliography{reference}

\appendices

{
\section{Proof of Linear Leakage Properties}

We provide mathematical proofs for the following two properties of the linear leakage primitive, which help to prove the Eq. \ref{recovery equation} in Section \ref{Preliminaries}.

\textbf{Property 1:} For $l$ in $\{1,2,\cdots,k\}$, considering $x_p$ is the $p^{th}$ smallest sample in terms of feature $h(x)$ and falls in the $l^{th}$ bin $[h_l, h_{l+1}]$ alone, $\nabla_{w_1(l+1)}L$ satisfies $\nabla_{w_1(l+1)}L=\frac{\partial L}{\partial y_{l+1}}\frac{\partial y_{(l+1)}}{\partial w_{1(l+1)}}=\sum\limits_{v=1}^{p}\frac{\partial L}{\partial y_{l+1}}x_v$.

\textbf{Proof:} According to the chain rule, we can calculate the gradient as $\nabla_{w_1(l+1)}L=\sum\limits_{v=1}^{k}\frac{\partial L}{\partial y_{l+1}}x_v$. We decompose this equation into two parts as $\sum\limits_{v=1}^{p}\frac{\partial L}{\partial y_{l+1}}x_v+\sum\limits_{v=p+1}^{k}\frac{\partial L}{\partial y_{l+1}}x_v$. For the second part, we have $y_{(l+1)}<0$ because all these $x_v$ are smaller than $x_p$ and can not activate bin $l$. Then according to the property of the ReLU function (zero gradients for negative values), $\frac{\partial L}{\partial y_{l+1}}=0$ for these values. This indicates that the second part is always zero and the property holds.

\textbf{Property 2:} By letting the row vectors of $w_2$ identical, we have $\frac{\partial L}{\partial y_{l+1}}=\frac{\partial L}{\partial y_{l}}$.

\textbf{Proof:} According to the chain rule, we have $\frac{\partial L}{\partial y_{l+1}}= \sum\limits_{i=1}^o\frac{\partial L}{\partial z_{i}}\frac{\partial z_{i}}{\partial y_{l+1}}$. Because all the row vectors of $w_2$ are identical, $\frac{\partial z_{i}}{\partial y_{l+1}}=\frac{\partial z_{i}}{\partial y_{l}}$ for all $i \in \{1,2,\cdots,o\}$. Then we take it back and have $\sum\limits_{i=1}^o\frac{\partial L}{\partial z_{i}}\frac{\partial z_{i}}{\partial y_{l+1}}=\sum\limits_{i=1}^o\frac{\partial L}{\partial z_{i}}\frac{\partial z_{i}}{\partial y_{l}}=\frac{\partial L}{\partial y_{l}}$.
}

\section{Experiment Datasets}

The FMNIST dataset consists of a training set of 60,000 samples and a test set of 10,000 samples. Each sample is a $28\times 28$ grayscale image, associated with a label from 10 classes, including T-shirts, trousers, pullovers, dresses, coats, sandals, shirts, sneakers, bags, and ankle boots. 
HMNIST is a medical dataset that contains 5000 images for 8 types of skin cancers. Each sample is a $28\times 28$ grayscale image, associated with a label from 8 classes of cancers. 
The CIFAR-10 dataset consists of 60,000 $32\times 32$ color images in 10 classes, with 50,000 training images and 10,000 test images. Each image is from one of the ten classes, including airplanes, automobiles, birds, cats, deer, dogs, frogs, horses, ships, and trucks. 
The TinyImageNet dataset contains 120000 images of 200 classes (600 for each class) sized to 64×64 colored images. Each class has 500 training images, 50 validation images, and 50 test images. 
{ The CelebFaces Attributes (CelebA) dataset contains 202,599 face images from 10,177 celebrities of size 178×218, each annotated with 40 binary labels indicating facial attributes like hair color, gender, and age. The training set consists of 162770 images, the test set consists of 19867 images and the validation set consists of 19962 images. The ImageNette dataset is a subset of the super large ImageNet dataset that contains 10 out of 1000 classes in the original ImageNet dataset, including Tench (a type of fish), English springer (dog breed), Cassette player, Chain saw, Church, French horn, Garbage truck, Gas pump, Golf ball, and Parachute. The dataset contains 13000 images of pixel size 320×320.}

\begin{table}[t]
    \centering
    \caption{ Attack performance on the FashionMNIST dataset with and without VAE regulation.}
    \small
    \centering
    \begin{tabular}{lccccc}
        \toprule
         \multirow{2}{*}{Batch}  & \multicolumn{2}{c}{Without VAE} & \multicolumn{2}{c}{VAE Regulated}\\
         & Rate & PSNR & Rate & PSNR \\
        \midrule
        64 &  0.9402 &  39.4908 &  0.9625 &  28.0435 \\
        128 &  0.9057 &  38.6366 &  0.9107 &  27.8447 \\
        256 &  0.7882 &  34.2213 &  0.8775 &  27.3211 \\
        512 &  0.6613 &  33.9857 &  0.8307 &  26.4764 \\
        \bottomrule
    \end{tabular}
    \vspace{-0.1in}
    \label{tab:fmnist-vae-regulated}
\end{table} 

\begin{figure}[t]
\centering
    \subfigure[ LSR distribution of FMNIST dataset without regulation.]{\includegraphics[width=0.235\textwidth]{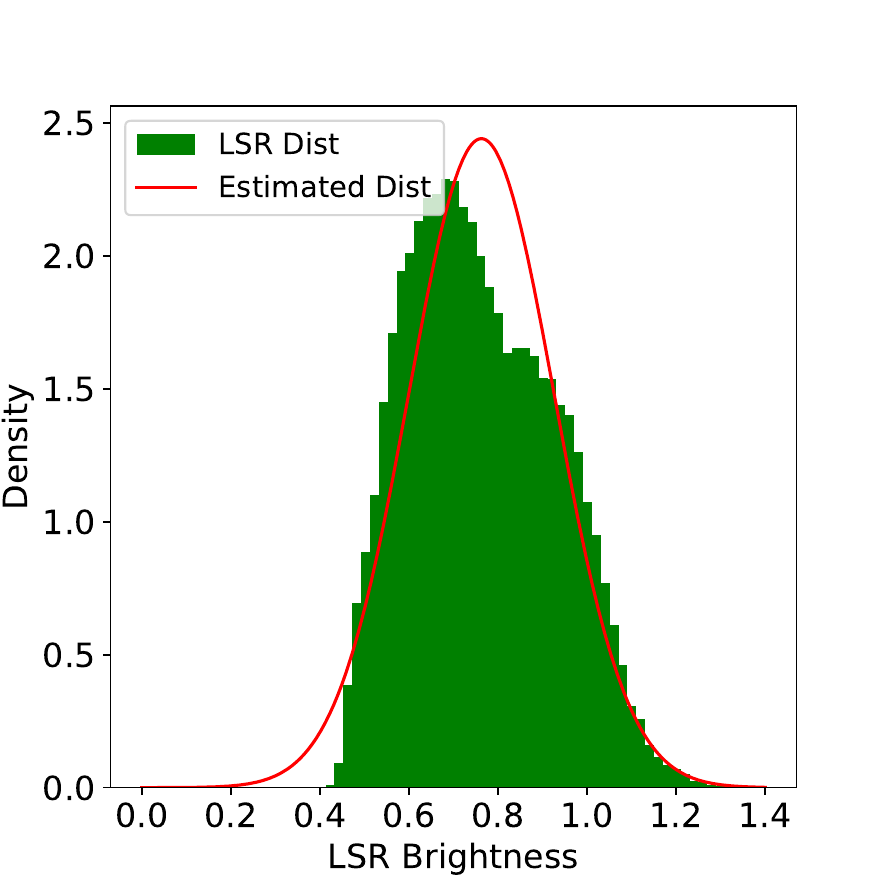}}
    \subfigure[ LSR distribution of FMNIST dataset with VAE regulation method.]{\includegraphics[width=0.235\textwidth]{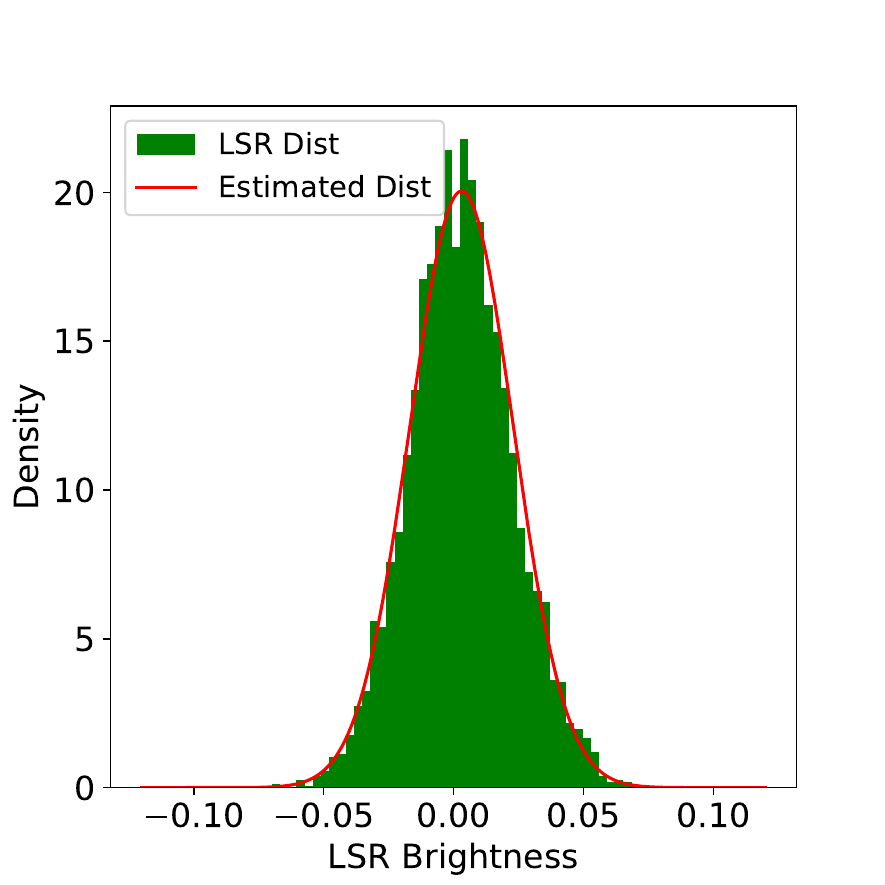}}
    \caption{ The LSR distribution of the FMNIST dataset before and after using the VAE regulation method. }
    \label{fig: fmnist-vae} 
    \vspace{-0.15in}
\end{figure}

\section{Regulating LSR Distribution}

In this section, we evaluate the \sysname's attack performance when the LSR distribution is regulated to certain distributions. We clarify that the attacker controls this regulation process during the attack preparation phase. More specifically, the attacker can regulate the surrogate autoencoder's local training process to ensure that the LSR distribution of the auxiliary dataset follows pre-defined distributions when the surrogate model is converged. We assume the attacker uses the widely used variational autoencoder (VAE) as the regulation method. We evaluated the regulation performance on the FashionMNIST dataset as without regulation the LSR distribution of the FashionMNIST dataset is biased and not similar to the standard Gaussian distribution. In Figure. \ref{fig: fmnist-vae}, we demonstrate the LSR distribution of the dataset before and after the VAE regulation. From the figure, we can clearly observe that the LSR distribution of the FashionMNIST dataset becomes an almost perfect Gaussian distribution after the VAE is used. This makes the statistical estimation much easier. We further demonstrate attack performance in Table. \ref{tab:fmnist-vae-regulated}. We find that using VAE will make the PSNR score lower but remain in a decent range. The main benefit of using VAE is that it makes the LSR distribution shapes better and this contributed to better reconstruction rates, which is validated by the reconstruction rate performance in Table. \ref{tab:fmnist-vae-regulated}. 

\section{Batched Reconstruction Examples}
\begin{figure*}[p]
\centering
    \subfigure[A batch of 64 images from the CIFAR-10 dataset.]{\includegraphics[width=0.46\textwidth]{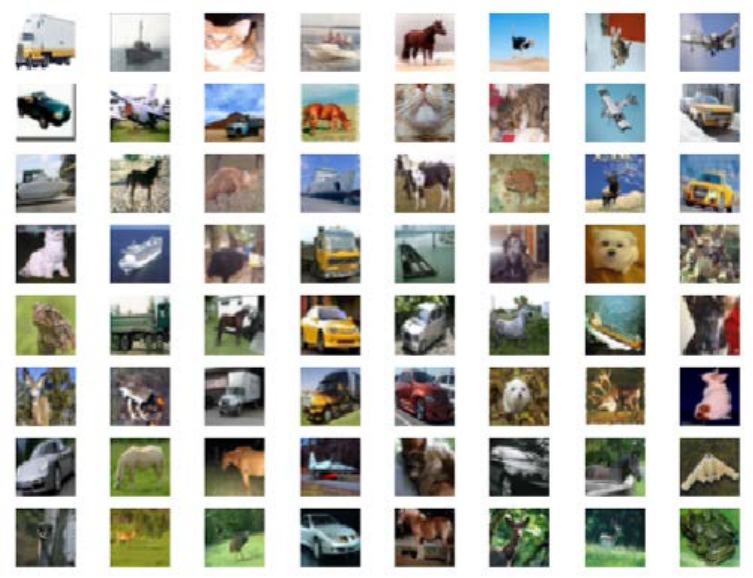}}
    \subfigure[The reconstructed images for the CIFAR-10 input batch.]{\includegraphics[width=0.47\textwidth]{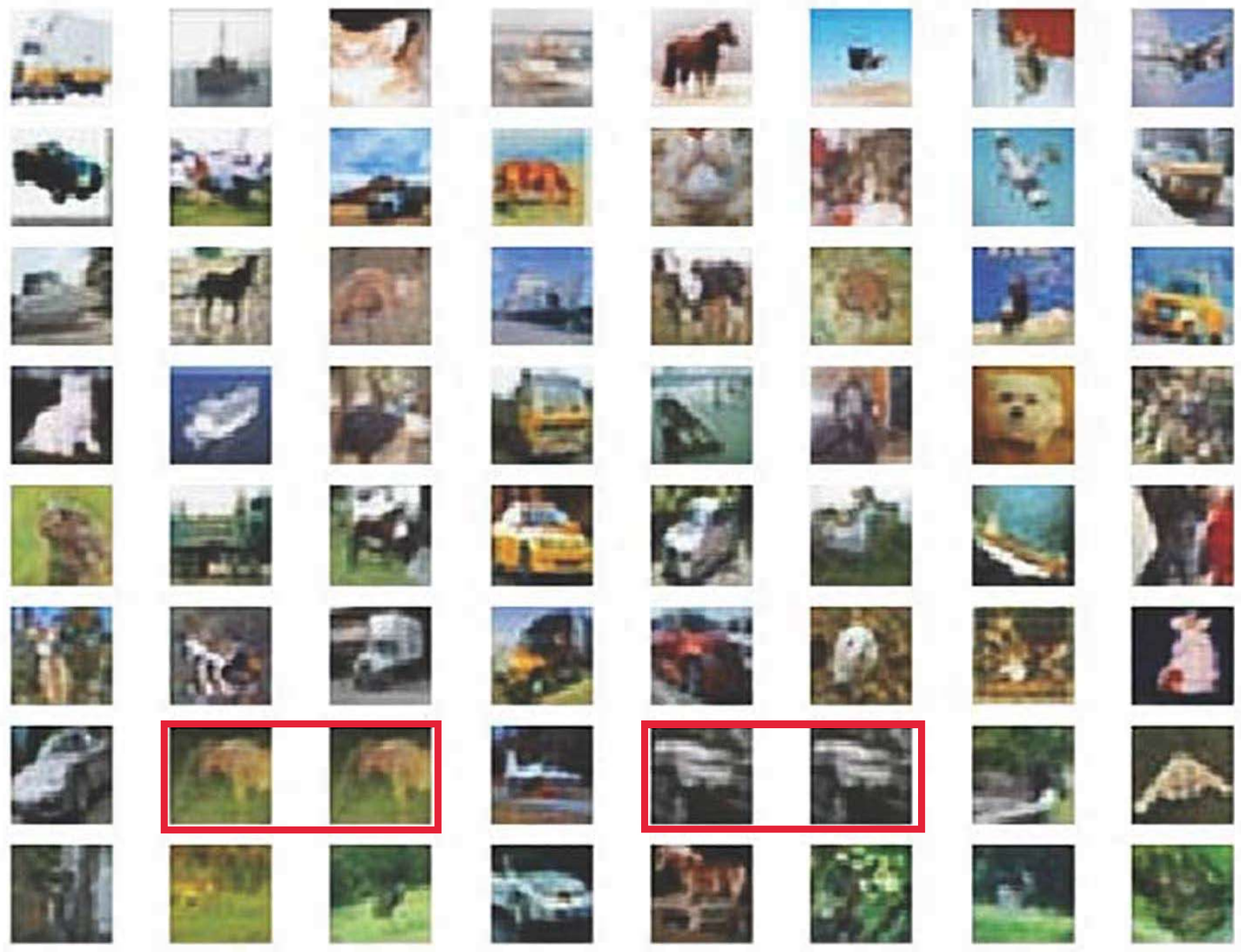}}
    \vspace{-0.12in}
    \caption{The comparison between the original images and the reconstructed images with batch size 64 on CIFAR-10. 
    }
    \label{fig: Batch-example-cifar} 
    \vspace{-10pt}
\end{figure*}

\begin{figure*}[t]
\centering
    \subfigure[A batch of 64 images from the FMNIST dataset.]{\includegraphics[width=0.46\textwidth]{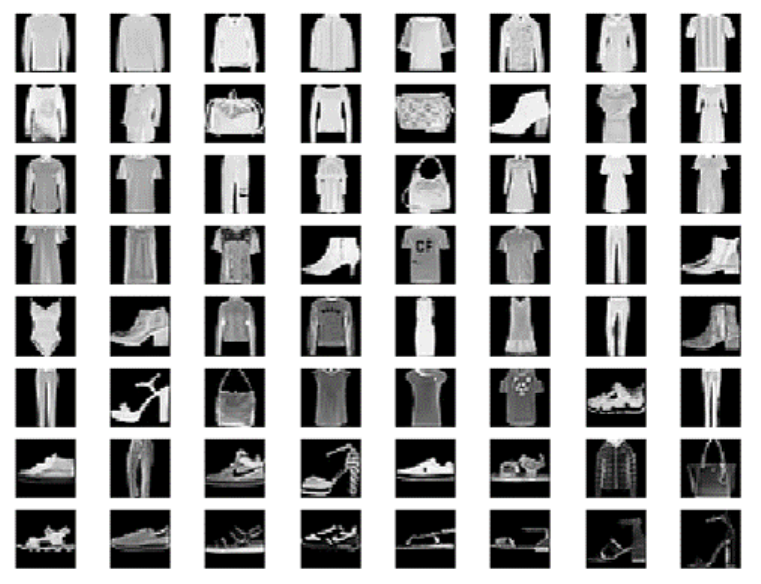}}
    \subfigure[The reconstructed images for the FMNIST input batch.]{\includegraphics[width=0.46\textwidth]{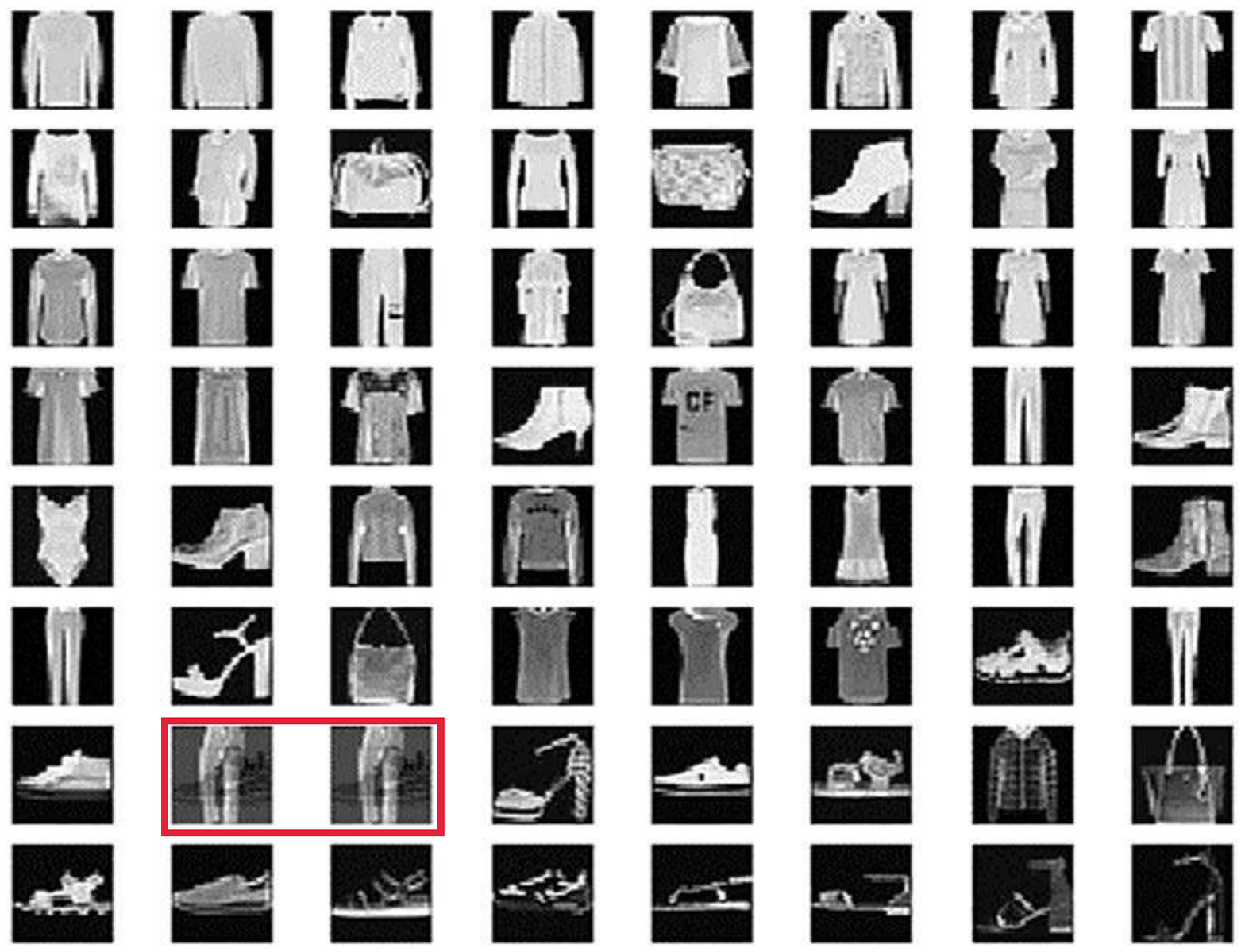}}
    \vspace{-0.12in}
    \caption{The comparison between the original images and the reconstructed images with batch size 64 on FMNIST. }
    \label{fig: Batch-example-fmnist} 
    \vspace{-10pt}
\end{figure*}

\begin{figure*}[p]
\centering
    \subfigure[A batch of 64 images from the HMNIST dataset.]{\includegraphics[width=0.46\textwidth]{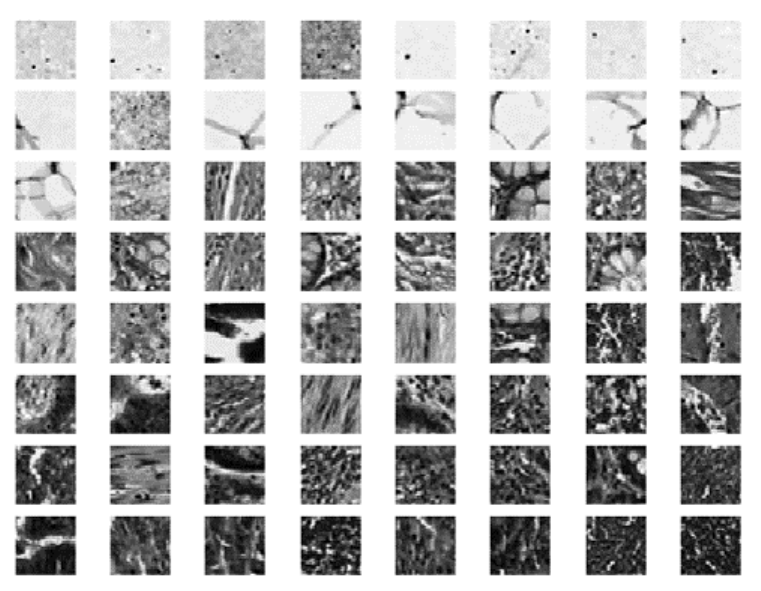}}
    \subfigure[The reconstructed images for the HMNIST input batch.]{\includegraphics[width=0.47\textwidth]{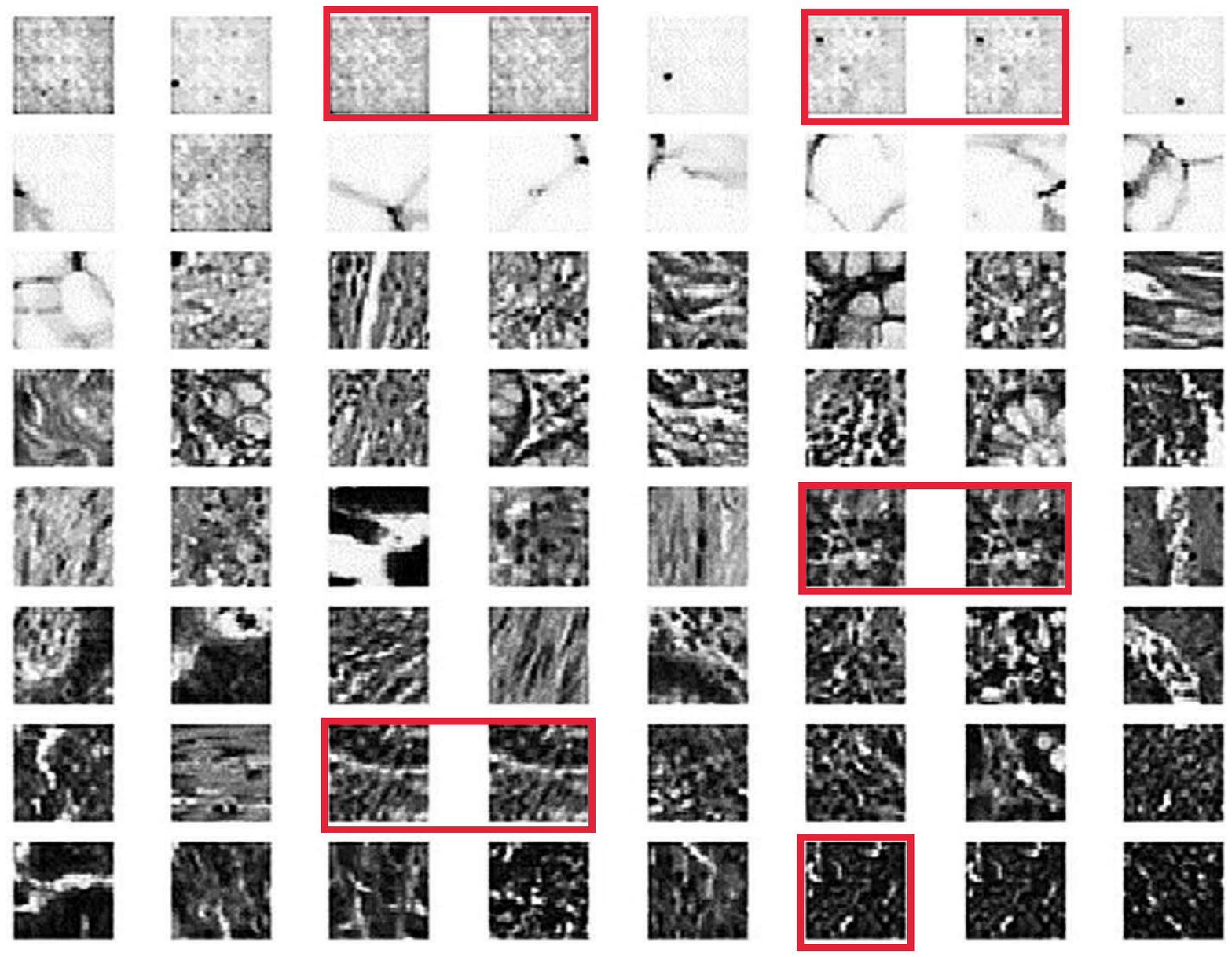}}
    \vspace{-0.12in}
    \caption{The comparison between the original images and the reconstructed images with batch size 64 on HMNIST. }
    \label{fig: Batch-example-hmnist} 
    \vspace{-10pt}
\end{figure*}

\begin{figure*}[h]
\centering
    \subfigure[A batch of 64 images from the TinyImageNet dataset.]{\includegraphics[width=0.45\textwidth]{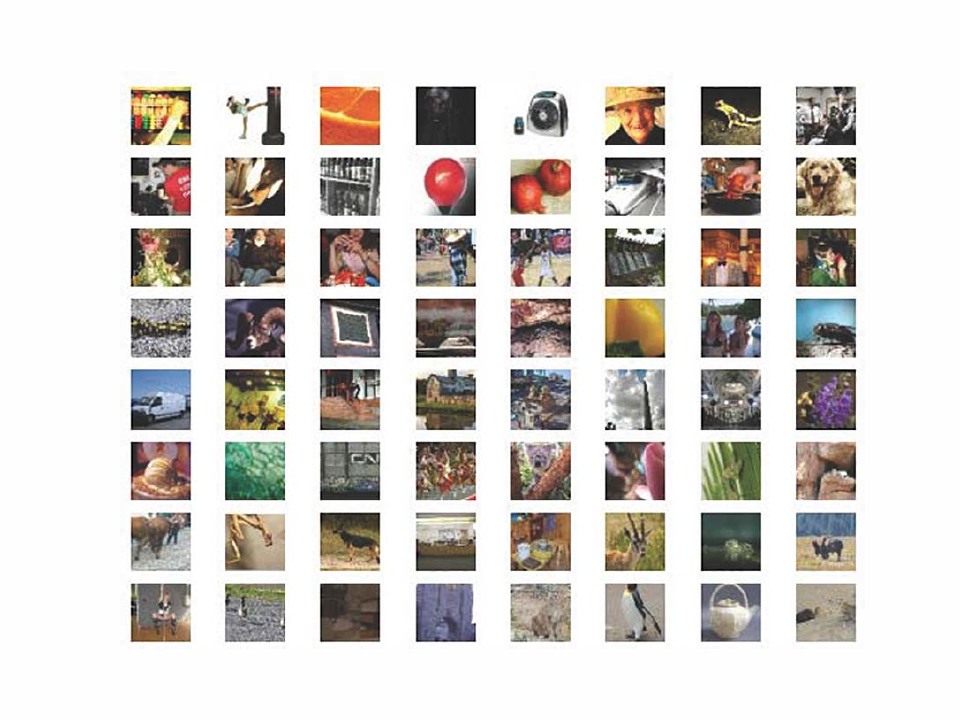}}
    \hspace{8pt}
    \subfigure[The reconstructed images for the TinyImageNet input batch.]{\includegraphics[width=0.45\textwidth]{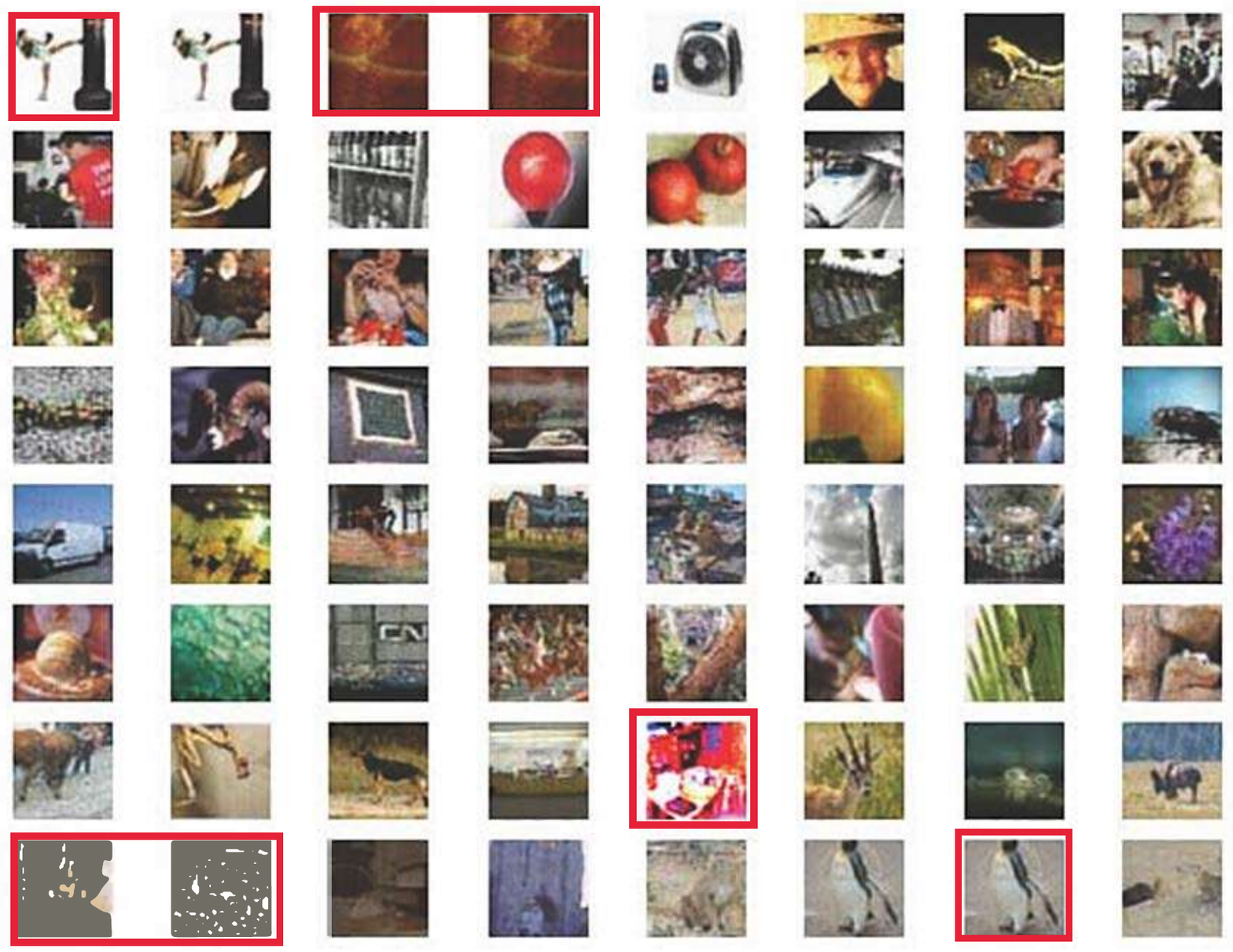}}
    \vspace{-0.12in}
    \caption{The comparison between the original images and the reconstructed images with batch size 64 on TinyImageNet. }
    \label{fig: Batch-example-tiny} 
   \vspace{-10pt}
\end{figure*}

\begin{figure*}[h]
\centering
    \subfigure[{ A batch of 64 images from the ImageNette dataset.}]{\includegraphics[width=0.45\textwidth]{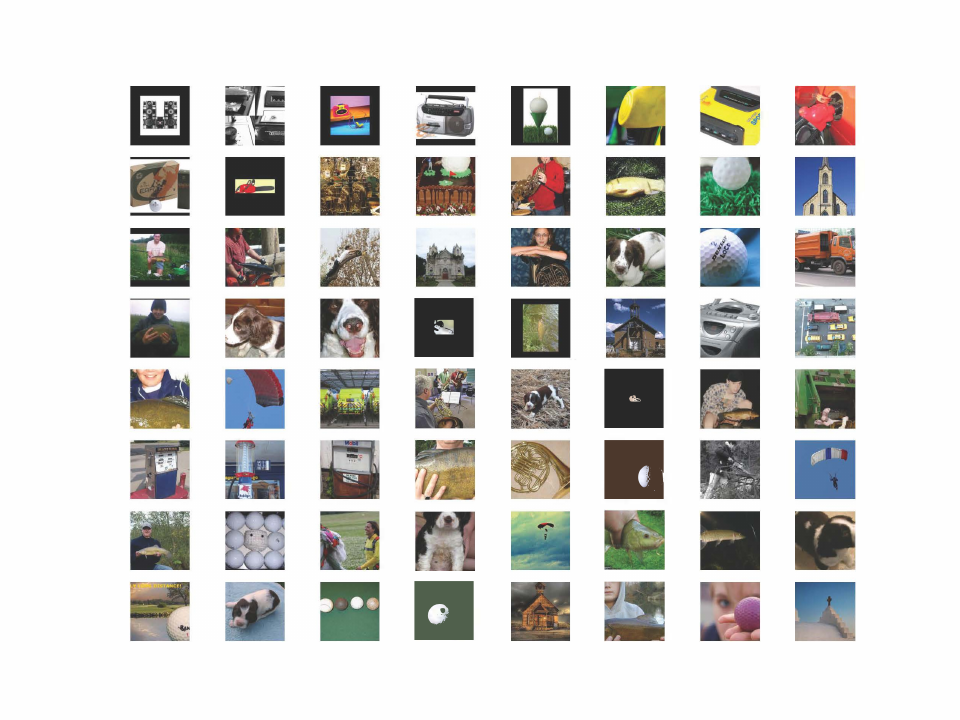}}
    \hspace{8pt}
    \subfigure[{ The reconstructed images for the ImageNette input batch.}]{\includegraphics[width=0.45\textwidth]{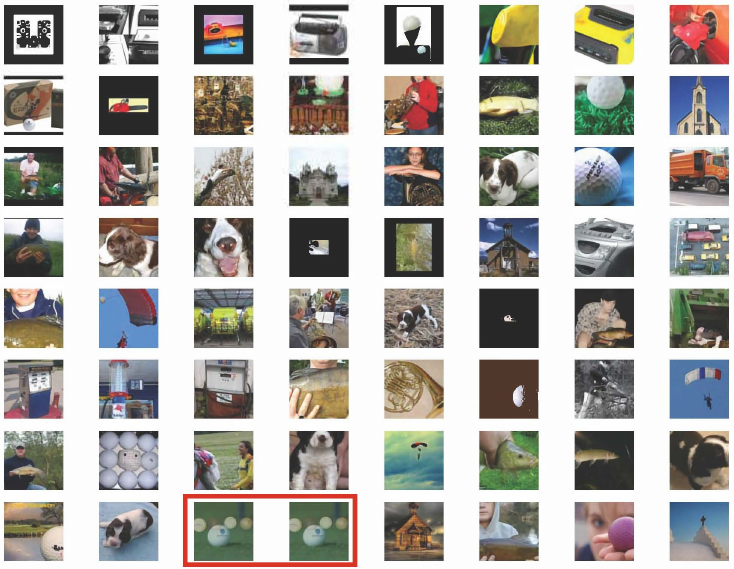}}
    \vspace{-0.12in}
    \caption{{ The comparison between the original images and the reconstructed images with batch size 64 on ImageNette.} }
    \label{fig: Batch-example-imagenet} 
   \vspace{-10pt}
\end{figure*}

\begin{figure*}[h]
\centering
    \subfigure[{ A batch of 64 images from the CelebA dataset.}]{\includegraphics[width=0.45\textwidth]{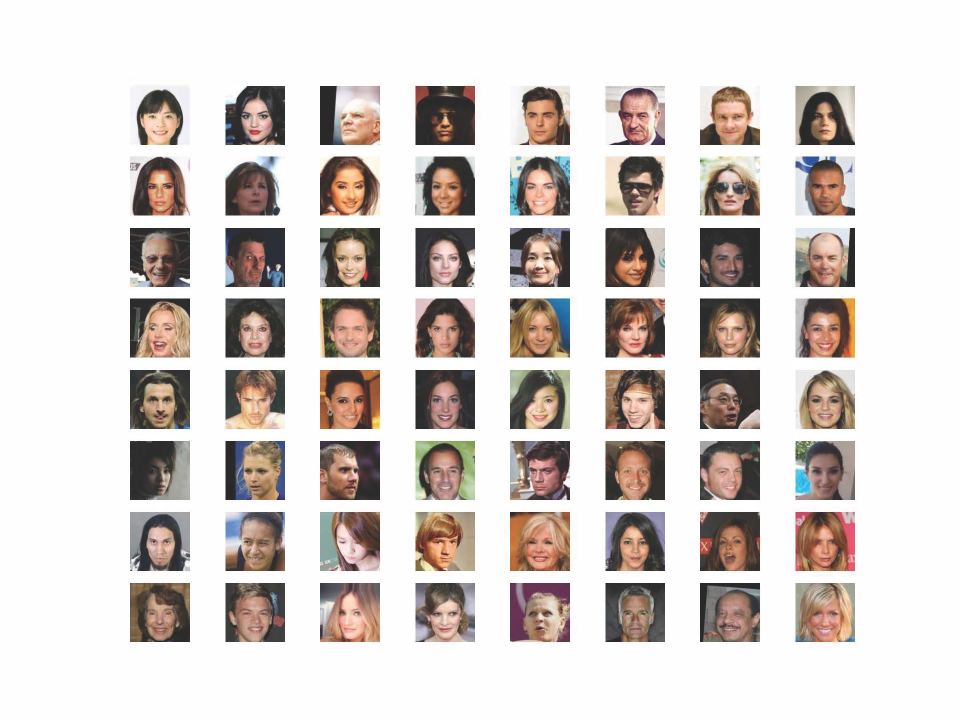}}
    \hspace{8pt}
    \subfigure[{ The reconstructed images for the CelebA input batch.}]{\includegraphics[width=0.45\textwidth]{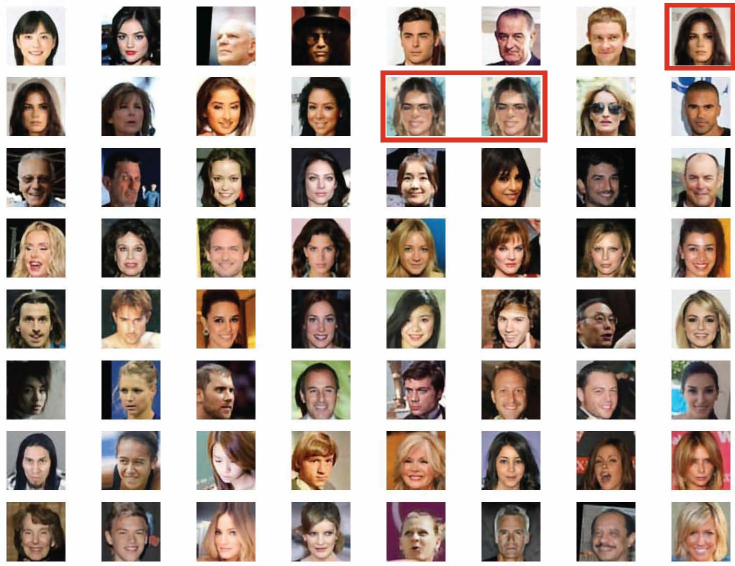}}
    \vspace{-0.12in}
    \caption{{ The comparison between the original images and the reconstructed images with batch size 64 on CelebA.} }
    \label{fig: Batch-example-celeba} 
   \vspace{-10pt}
\end{figure*}

We plot six randomly selected reconstructed batches with batch size 64 from the CIFAR-10, FMNIST, HMNIST, TinyImageNet, { ImageNette, and CelebA} datasets in Figure \ref{fig: Batch-example-cifar}, Figure \ref{fig: Batch-example-fmnist}, Figure \ref{fig: Batch-example-hmnist}, Figure \ref{fig: Batch-example-tiny}, { Figure \ref{fig: Batch-example-imagenet}, and Figure \ref{fig: Batch-example-celeba}} respectively. 

For each figure, we plot the original images in one subfigure and the reconstructed ones in the other. We can observe excellent reconstruction rates and good PSNR scores on all four batches. For the four datasets, 60, 62, 55, 57, 62, and 61 out of 64 images are successfully reconstructed respectively. 
The successfully reconstructed images are very clear and visually identical to the original ones, and a curious attacker can easily tell all the meaningful contents from these reconstructed images. 
We also mark the reconstruction failure images in red rectangles. We observe that the failure images usually collide with their neighborhoods and are mixed with each other during the reconstruction process (i.e. they fall in the same reconstruction bin). For these collided pairs, usually one sample instead of two is repeatedly reconstructed. There are also a few reconstruction failure samples that completely collapse and no useful information can be extracted from them.

\section{Reconstruction Statistics}

\begin{figure}[h]
    \centering
    \subfigure[Batch size = 128.]{\includegraphics[width=0.24\textwidth]{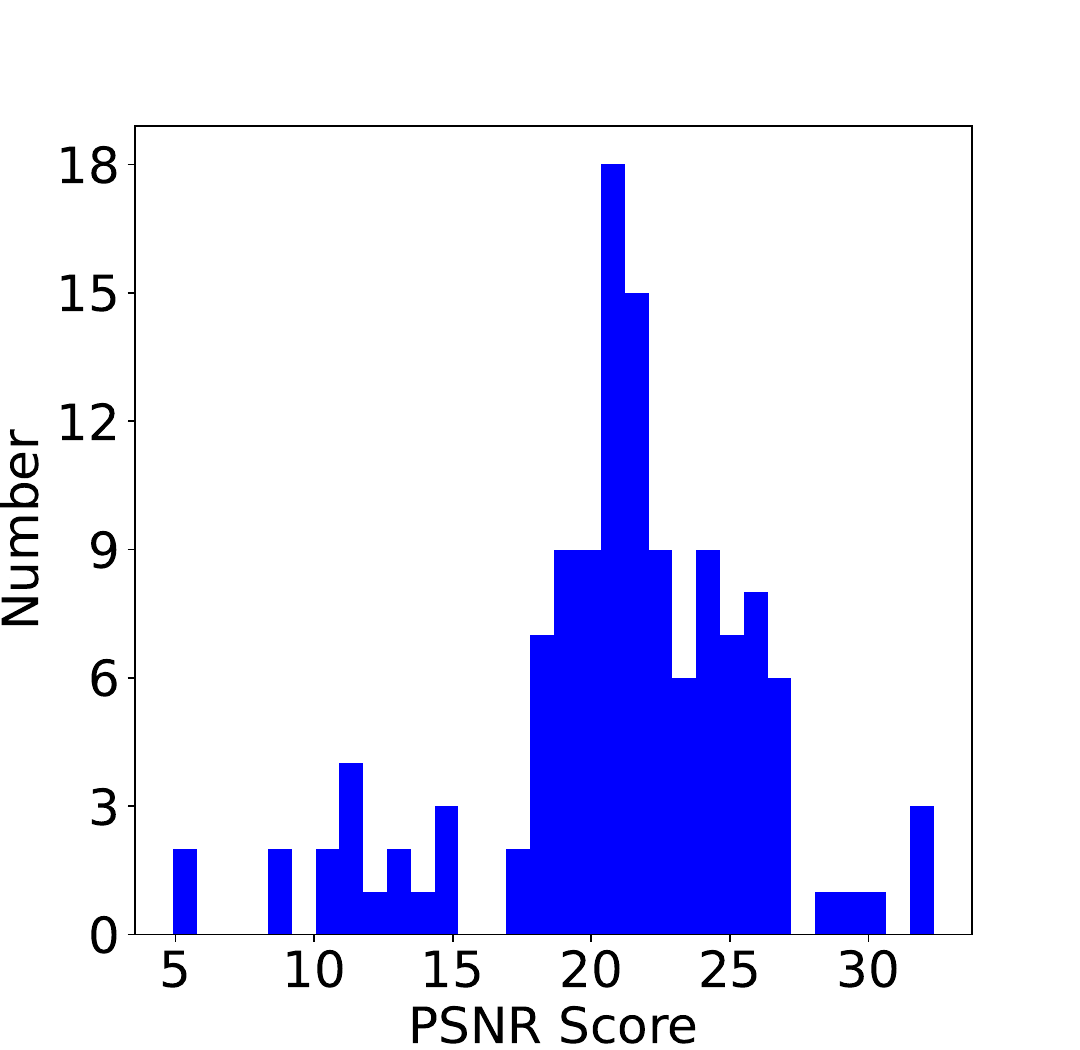}}
    \subfigure[Batch size = 256.]{\includegraphics[width=0.24\textwidth]{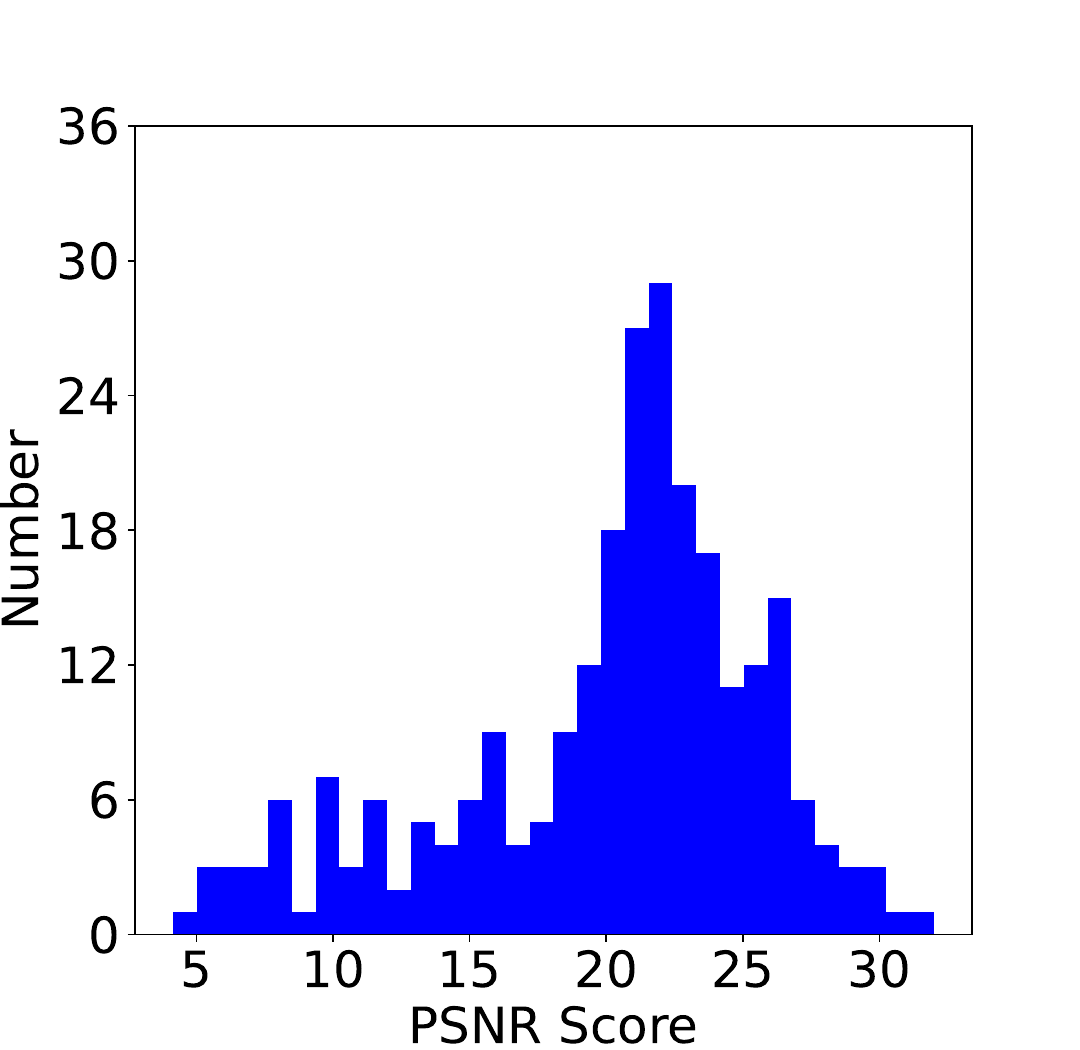}}
    \subfigure[Batch size = 512.]{\includegraphics[width=0.24\textwidth]{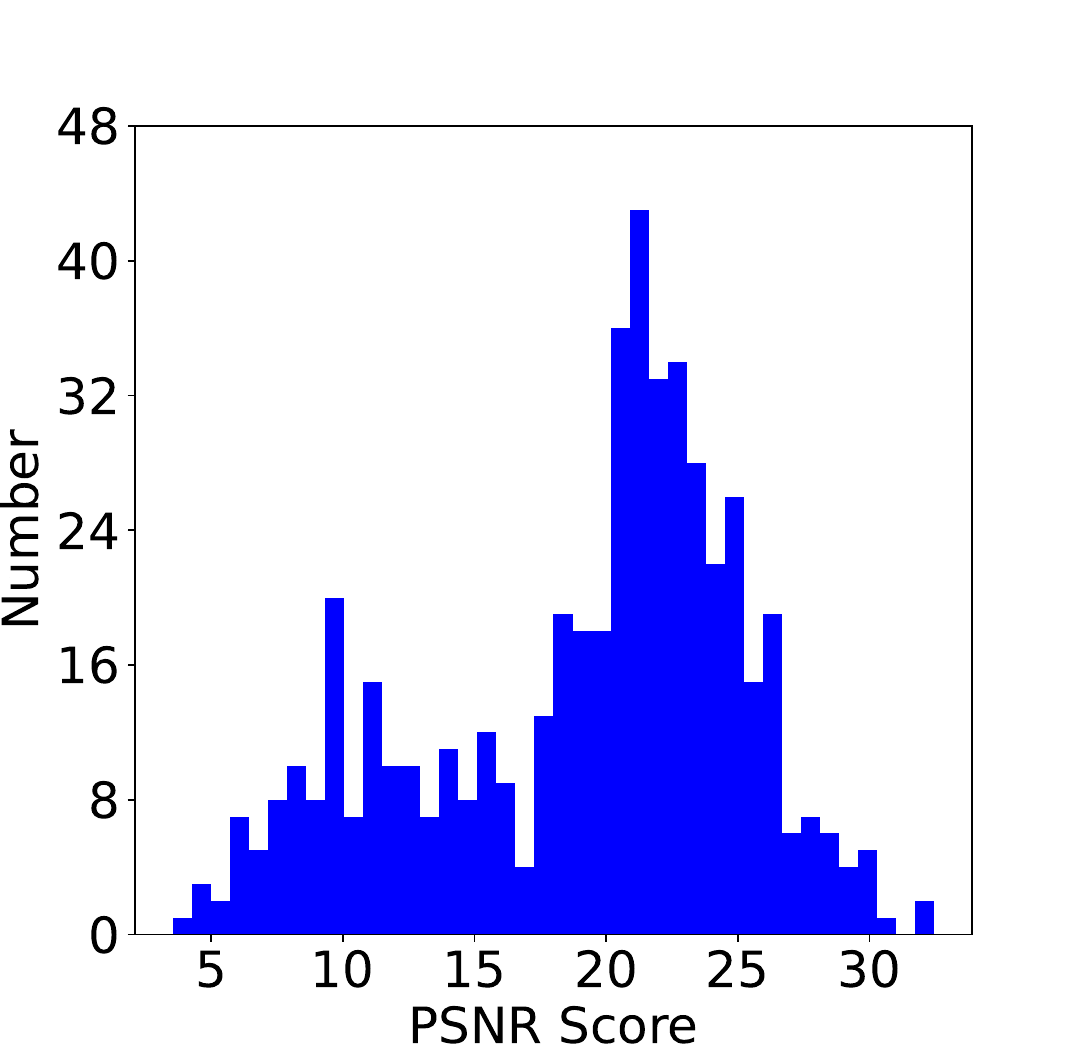}}
    \subfigure[Batch size = 1024.]{\includegraphics[width=0.24\textwidth]{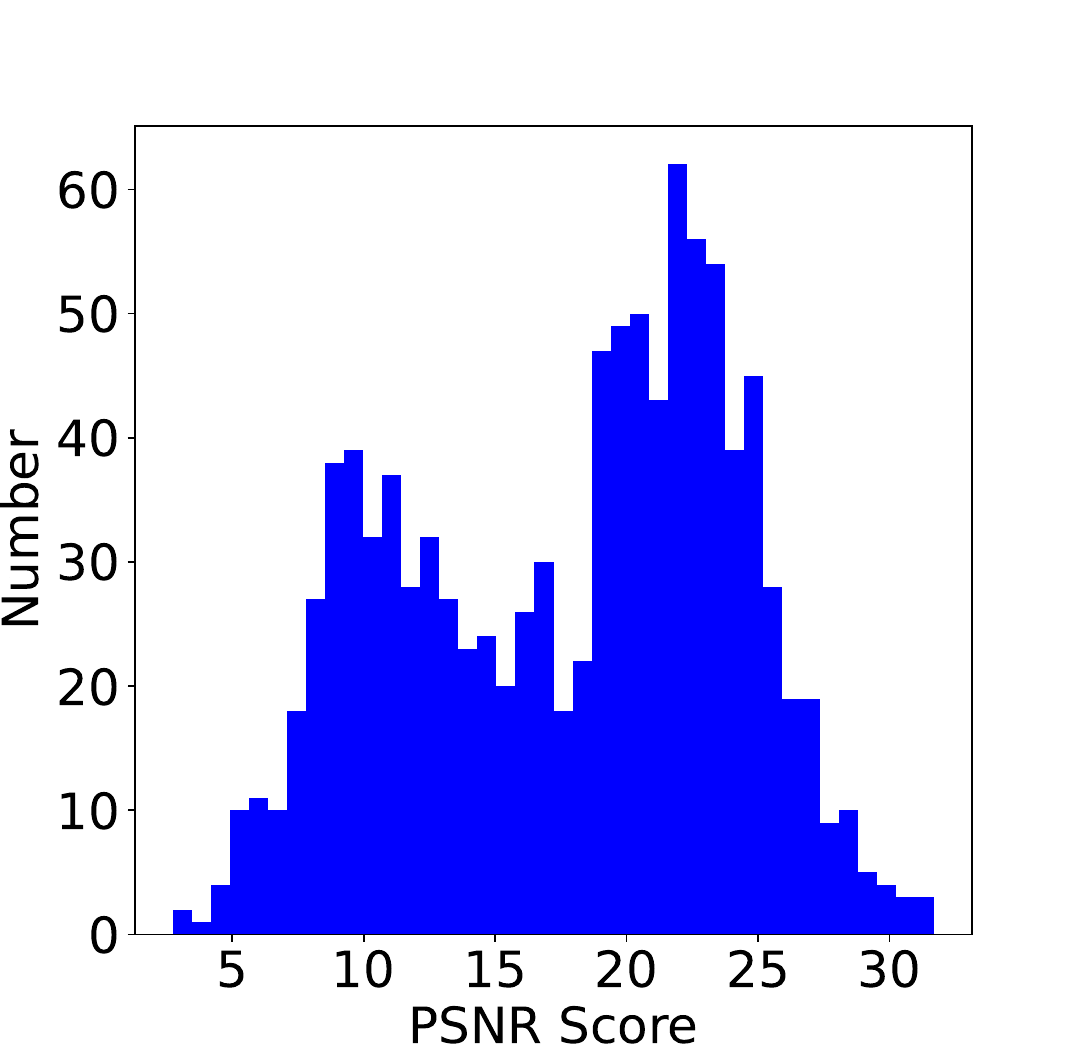}}
    \caption{The PSNR distribution of different reconstructed input batches for the TinyImageNet dataset.}
    \vspace{-0.15in}
    \label{fig: PSNR-dist} 
\end{figure}

In Figure \ref{fig: PSNR-dist}, we plot the PSNR distribution of four randomly selected batches with sizes \{128, 256, 512, 1024\} reconstructed by CNN Net on the TinyImageNet dataset. We can clearly find that with larger batch sizes the PSNR distribution becomes worse and more reconstructed samples fall below the threshold $th=18$. We can also observe that the samples approximately follow a bi-modal distribution. This is because when the batch size becomes larger, more samples fall into the same reconstruction bins (in total 1024 for CNN Net), resulting in more collisions and reconstruction failures. The collision and failed samples form one model in the distribution and the successfully reconstructed samples form the other.

\begin{figure}[h]
    \centering
    \subfigure[CIFAR-10.]{\includegraphics[width=0.24\textwidth]{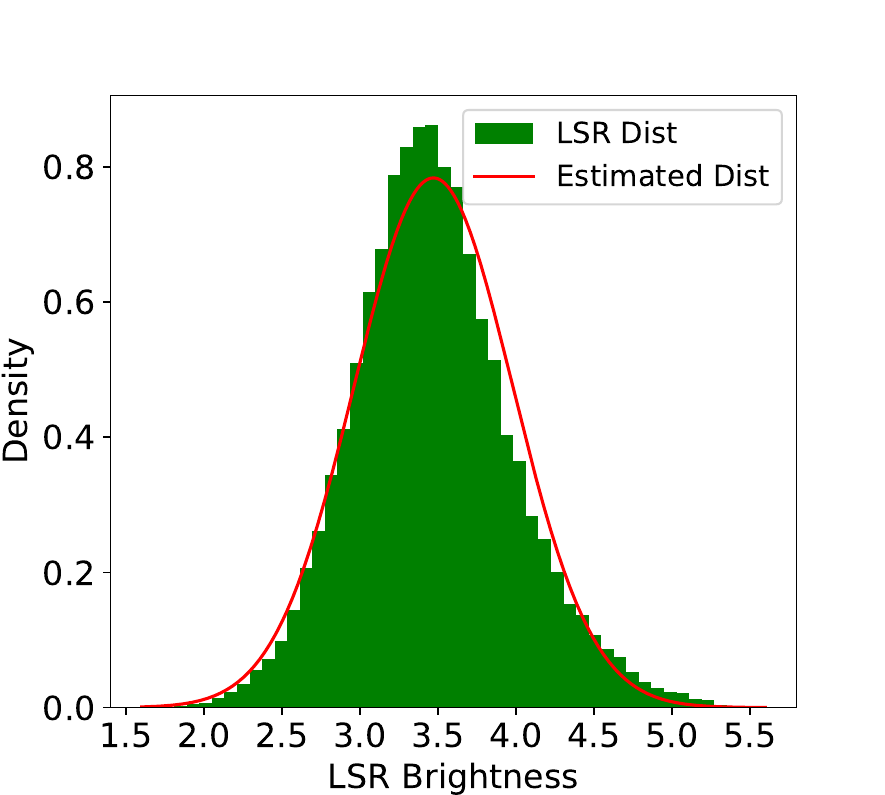}}
    \subfigure[FMNIST.]{\includegraphics[width=0.24\textwidth]{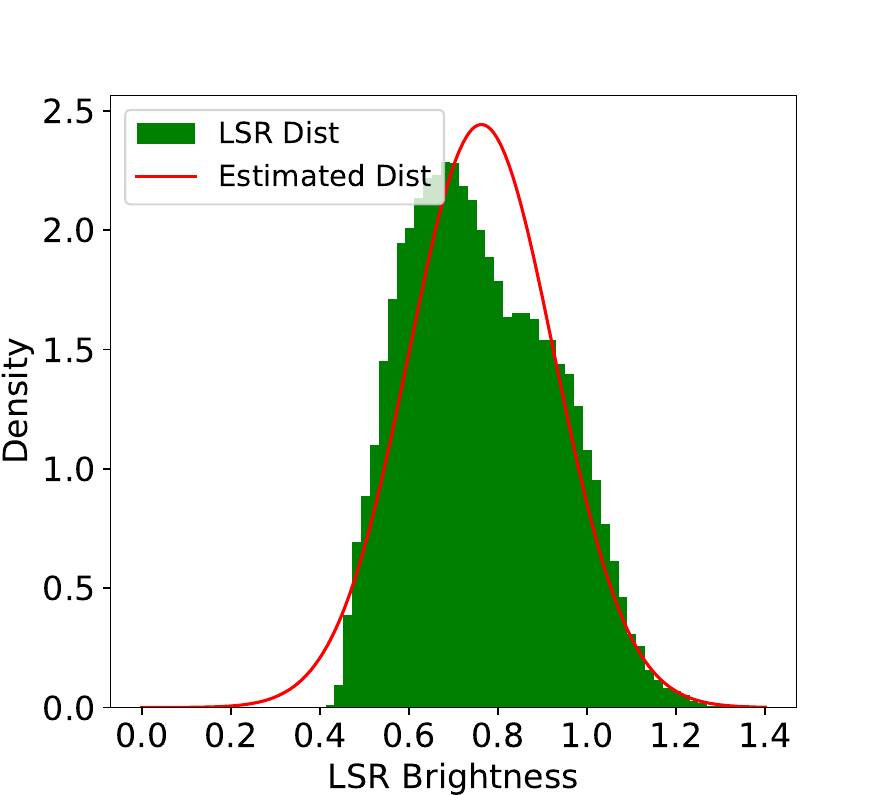}}
    \subfigure[HMNIST.]{\includegraphics[width=0.24\textwidth]{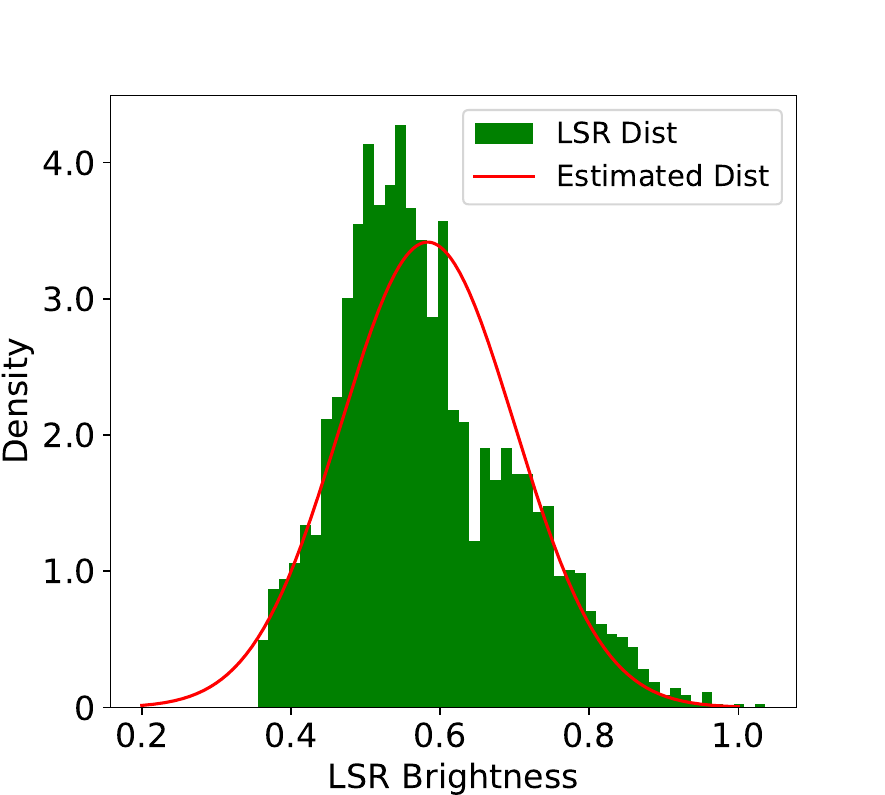}}
    \subfigure[TinyImageNet.]{\includegraphics[width=0.24\textwidth]{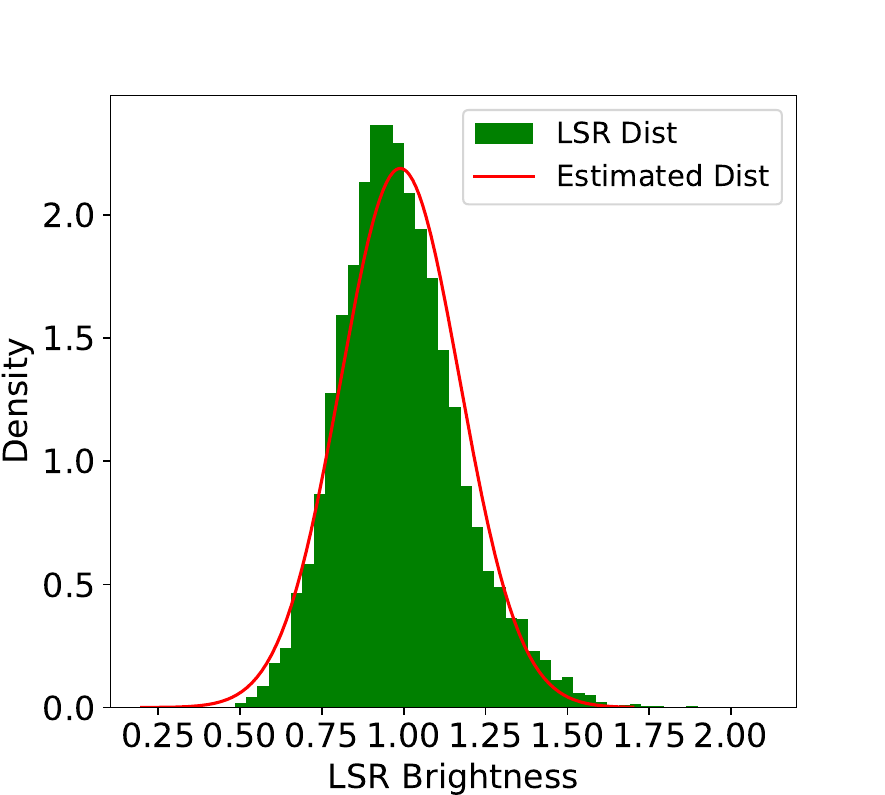}}
    \caption{The LSR brightness distribution of different datasets.}
    \label{fig: Data-dist-and-estimation} 
    \vspace{-0.15in}
\end{figure}

In Figure \ref{fig: Data-dist-and-estimation}, we show the LSR brightness distribution on different datasets, as brightness is the unique feature we used to craft the ``linear leakage'' module.
We observe that the LSR distribution of the CIFAR-10, and TinyImageNet datasets are highly similar to the Gaussian distribution and are easy to estimate. However, we also notice imperfect estimations including high peaks and bias for the FMNIST, and HMNIST datasets. These imperfect distribution estimations can finally degrade the attack performance, although not significant in our experiments. To fix them, we consider the attacker can use the variational autoencoder (VAE) to replace the current traditional autoencoder, which can regulate the latent space distribution to standard Gaussian distribution.

\newpage

\newpage

\end{document}